\documentclass{article}

\usepackage{wrapfig}
\usepackage{hyperref}
\usepackage{url}
\usepackage{microtype}
\usepackage{graphicx}
\usepackage{subfigure}
\usepackage{booktabs} % for professional tables
\usepackage{xparse} 
\usepackage[utf8]{inputenc}
\usepackage[linesnumbered,ruled]{algorithm2e}
\usepackage{textgreek}
\usepackage{tcolorbox}
\usepackage{amsmath}
\usepackage{mathtools}
\usepackage{amsthm}
\usepackage{multirow}
\usepackage{float}
\usepackage{hyperref}
\usepackage{dblfloatfix}
\usepackage[textsize=tiny]{todonotes}

\usepackage{xcolor}
\usepackage{color, colortbl,booktabs}
\definecolor{Gray}{gray}{0.9}

\newcounter{todocounter}

\usepackage{pifont}

\usepackage{adjustbox}
\NewDocumentCommand{\rot}{O{45} O{1em} m}{\makebox[#2][l]{\rotatebox{#1}{#3}}}%
\theoremstyle{plain}

\theoremstyle{definition}

\theoremstyle{remark}

\makeatletter
\newcommand{\removelatexerror}{\let\@latex@error\@gobble}
\makeatother

\usepackage{hyperref}

\usepackage[accepted]{icml2025}

\usepackage{amsmath}
\usepackage{amssymb}
\usepackage{mathtools}
\usepackage{amsthm}

\usepackage[capitalize,noabbrev]{cleveref}

\theoremstyle{plain}
\theoremstyle{definition}
\theoremstyle{remark}

\usepackage[textsize=tiny]{todonotes}

\definecolor{cyan}{RGB}{13, 255, 255}
\definecolor{hotpink}{RGB}{255,35,255}
\definecolor{navyblue}{RGB}{44,44,255}

\icmltitlerunning{Adaptive Sensitivity Analysis for Robust Augmentation against  Natural Corruptions in Image Segmentation}

\begin{document}

\twocolumn[
\icmltitle{Adaptive Sensitivity Analysis for Robust Augmentation against \\ Natural Corruptions in Image Segmentation}




\begin{icmlauthorlist}
\icmlauthor{Laura Zheng}{umd}
\icmlauthor{Wenjie Wei}{umd}
\icmlauthor{Tony Wu}{umd}
\icmlauthor{Jacob Clements}{umd}
\\
\icmlauthor{Shreelekha Revankar}{umd,cornell}
\icmlauthor{Andre Harrison}{army}
\icmlauthor{Yu Shen}{umd,adobe}
\icmlauthor{Ming C. Lin}{umd}
\end{icmlauthorlist}

\icmlaffiliation{umd}{Department of Computer
Science, University of Maryland, College Park, MD, USA}
\icmlaffiliation{cornell}{Department of Computer
Science, Cornell University, Ithaca, New York, USA}
\icmlaffiliation{army}{DEVCOM Army Research Laboratory, Adelphi, Maryland, USA}
\icmlaffiliation{adobe}{Adobe Research
San Jose, California, USA}

\icmlcorrespondingauthor{Laura Zheng}{lyzheng@umd.edu}

\icmlkeywords{Machine Learning, ICML, robustness, segmentation, computer vision, perturbation, data augmentation, sensitivity analysis}

\vskip 0.3in
]



\printAffiliationsAndNotice{}  % leave blank if no need to mention equal contribution

\begin{abstract}
Achieving robustness in image segmentation models is challenging due to the fine-grained nature of pixel-level classification. These models, which are crucial for many real-time perception applications, particularly struggle when faced with natural corruptions in the wild for autonomous systems. While sensitivity analysis can help us understand how input variables influence model outputs, its application to natural and uncontrollable corruptions in training data is computationally expensive. In this work, we present an adaptive, sensitivity-guided augmentation method to enhance robustness against natural corruptions. Our sensitivity analysis on average runs {\bf 10x} faster and requires about {\bf 200x} less storage than previous sensitivity analysis, enabling practical, on-the-fly estimation during training for a model-free augmentation policy. With minimal fine-tuning, our sensitivity-guided augmentation method achieves improved robustness on both real-world and synthetic datasets compared to state-of-the-art data augmentation techniques in image segmentation. Code implementation for this work can be found at: \href{https://github.com/laurayuzheng/SensAug}{https://github.com/laurayuzheng/SensAug}.
\end{abstract}

\section{Introduction}

Segmentation models are crucial in many applications, but they often face unpredictable and uncontrollable natural variations that can degrade their performance. For instance, mobile applications using segmentation for image reconstruction may encounter diverse noises due to varying environmental lighting, camera quality, and user handling.
Similarly, autonomous vehicles and outdoor robots will need to operate under a wide range of adverse weather conditions that are difficult to simulate accurately. Even in medical imaging, where conditions are more controlled, factors such as slight movements can introduce blur, affecting segmentation results. While poor-quality examples can sometimes be discarded and re-captured, such solutions are costly or impractical, especially in large-scale, ubiquitous use cases, with limited resources, or during real-time inference (e.g., failure in a navigating robot). Addressing these natural corruptions is challenging because they are hard to capture ahead of time in a predictable and controllable way, simulate or parameterize, yet they significantly impact model performance.

\begin{figure}
    \centering
    \includegraphics[width=1.0\linewidth]{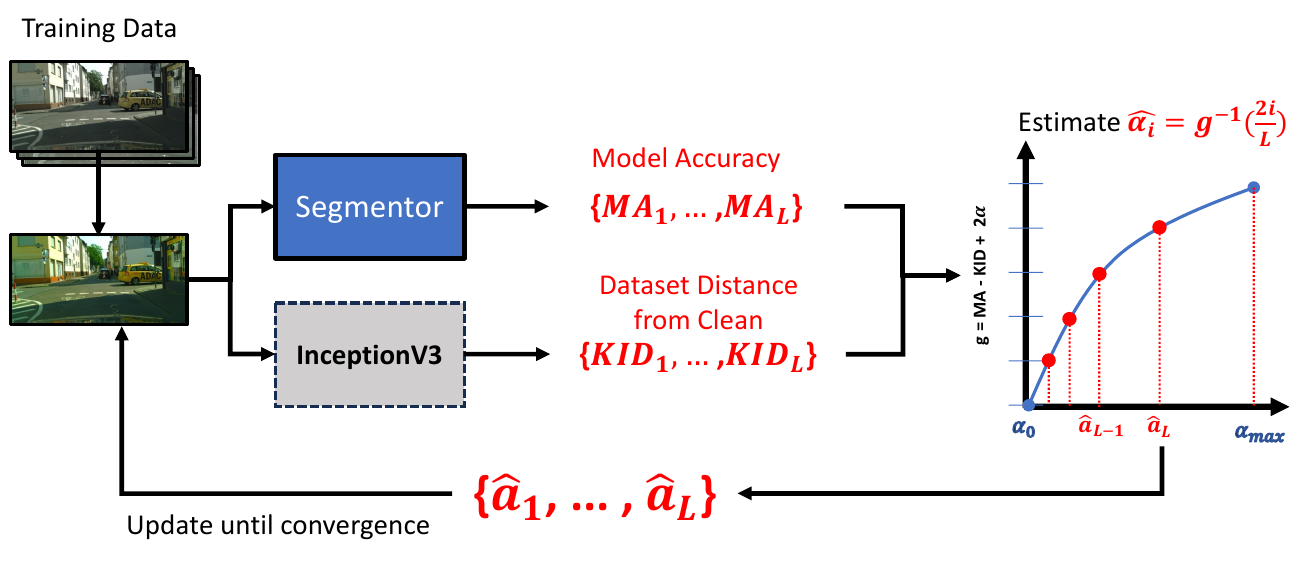}
    \vspace*{-2em}
    \caption{{\textbf{Overview of online sensitivity analysis.} We conduct {\em adaptive} sensitivity analysis using our fast analysis algorithm after a warmup period on clean data, then solve for $L$ discrete perturbation intensities, or levels, per perturbation type which the model is sensitive to. Finally, we augment training by sampling from the computed perturbation levels. Sampling weights are determined based off model performance on sensitive levels, where worse-performing levels are given higher probability of being sampled.}}
    \label{fig:overview}
    \vspace*{-1em}
\end{figure}

One common approach to enhance robustness against such corruptions is data augmentation, which artificially increases the diversity of training data by applying transformations to existing samples. While data augmentation is convenient and resource-efficient, its effectiveness depends on selecting the most beneficial augmentations.
Ideally, we should determine which augmentation a given model is most sensitive to and focus on those to improve performance---in other words, sensitivity analysis. However, traditional sensitivity analysis methods are computationally expensive and resource-intensive~\citep{yu_shen}, as shown in Table~\ref{tab:sa_complexity}, making them impractical for large-scale or real-time applications.
Existing methods like AutoAugment~\citep{autoaugment} and DeepAutoAugment attempt to optimize augmentation policies by training separate models, which adds significant overhead. Other state-of-the-art techniques rely on random augmentations~\citep{cubuk_2020, muller_2021, hendrycks_2020}, which are scalable but may not target the most impactful transformations.

In this paper, we propose a scalable, fast sensitivity-guided augmentation approach for robustifying segmentation models against natural corruptions, including those not explicitly involved during training. 
% While popular data augmentation techniques generally scale well, many require an external model trained offline to guide augmentation or rely on random augmentations. 
% Methods using sensitivity analysis (SA) provide improved performance over traditional augmentation techniques, but modeling the sensitivity of a model with augmentations regardless of differentiability can be resource-heavy on both runtime and storage.
% Our method uses the best of both worlds, offering a data augmentation technique that is practical and scalable, similarly to random augmentation techniques, but also retains the productivity of policy-based augmentation.
Our approach performs a lightweight, online sensitivity analysis during training to identify the geometric and photometric perturbations, shown to be effective as ``basis perturbations'' ~\citep{yu_shen}, to which the model is most sensitive. In contrast to ~\cite{yu_shen}, our sensitivity analysis is adaptive and significantly less resource intensive, allowing for practical implementation without the need for offline models or extensive computation. Figure~\ref{fig:overview} shows a high-level overview of our augmentation pipeline. Our method bridges the gap between the efficiency of random augmentation techniques and the effectiveness of policy-based augmentations guided by sensitivity analysis. Despite our focus on segmentation, our approach is general and can be applied to other tasks, architectures, or domains without significant modifications.

% Despite our focus on robustness for segmentation, our method is not bound to a particular task, architecture, or domain, and can be easily applied ad-hoc to any existing learning framework.

% Our method requires {\bf orders of magnitude {\em less} resources} compared to similar methods in dimensions of offline storage and computational runtime, which is due to the fact that we (1) do not require an additional network to systematically apply augmentations, and (2) do not require chained or complex online augmentations. 

% During training, our method conducts a {\em lightweight, online sensitivity analysis} to identify the geometric and photometric perturbations, shown to be effective as ``basis perturbations" ~\cite{yu_shen}, to which the model is most sensitive.
% In contrast to ~\cite{yu_shen}, we dynamically sample these perturbations and weight them based on the model's sensitivity at different intensity levels.

In experiments, we achieve up to a \textit{6.20}\% relative mIoU improvement in snowy weather and up to a \textit{3.85}\% relative mIoU improvement in rainy weather compared to the next-best method in zero-shot adverse weather evaluation on state-of-the-art architectures. We also show improvements on synthetic benchmarks and increased data efficiency (up to ${\bf 200x}$) compared to other augmentation methods as the size of the training set changes.

Our contributions are summarized as follows: 
\begin{enumerate}
    \item An efficient {\em adaptive} sensitivity analysis method for {\em online model evaluation} that iteratively approximates model sensitivity curves for significant speedup;
    \item A comprehensive, sensitivity-guided framework that systematically improves the robustness of learning-based segmentation models;
    \item Evaluation and analysis of our method on {\em unseen} synthetically perturbed samples, {\em naturally corrupted} samples, and ablated contributing factors to robustification. 
\end{enumerate}

\section{Related Works}

\textbf{Robustification Against Natural Corruptions}.
The effect of natural corruptions on deep learning tasks is a well-explored problem, especially in image classification. Currently, image classification has a robust suite of benchmarks, including evaluation on both synthetic and natural corruptions~\citep{ hendrycks_2020,yi2021benchmarking,Dong_2020_CVPR}.  Many works study correlations between image corruptions and various factors~\citep{mintun2021_augmentation_corruption,hendrycks2017a}. 
Additionally, a popular approach to increasing robustness in the general case is through targeted adversarial training~\citep{xu2021ddcat, shu_2021}.
Several approaches target model architecture~\citep{schneider2020_covariate_robustness,Saikia_2021_ICCV, myronenko_braintumor_seg_2020}. Other approaches achieve robustness to natural corruptions via the data pipeline. Data augmentations are a popular method for increasing out-of-distribution robustness and many have now become standard practice~\citep{geirhos2018imagenettrained,rusak2020_simple_robustness}.
Hendrycks et al. highlight that existing methods for generalization may not be consistently effective, emphasizing the need for robustness through addressing multiple distribution shifts~\citep{Hendrycks_2021_ICCV}.
In our work, we focus on studying and improving robustness in the context of semantic segmentation models due to natural corruptions using insights from previous work. 
% Firstly, we consider the robustness benchmark from~\cite{Kamann_2020_CVPR} that robustness in segmentation models differ from the traditional image classification setting used in most robustness approaches, in that the clean accuracy of segmentation models scale directly with robustness. Secondly, we also consider from~\cite{mintun2021_augmentation_corruption} that augmentations visually correlate with the ``type" of robustness it provides. 
Among findings from other works, we distinguish that our work focuses on improving natural corruption robustness in semantic segmentation, a common computer vision task with unique properties. 

\textbf{Data Augmentation Techniques}. 
Data augmentation methods generate variants of the original training data to improve model generalization capabilities. These variants do not change the inherent semantic meaning of the image, and transformed images are typically still recognizable by humans.
Within data augmentation methods, CutMix and AugMix are widely-used augmentation techniques that augment images by mixing variants of the same image~\citep{hendrycks_2020,yun_2019}.
Conversely,~\cite{franchi2021robustsemanticsegmentationsuperpixelmix} introduces segmentation-specific augmentation approaches which utilize superpixels, or clusters of similar pixels, to maintain semantic object information.
Other data augmentation methods have utilized augmentation policies based on neural networks to select productive augmentations \citep{Olsson_2021_WACV,autoaugment,yu2022deepaa},
while other works have explored data augmentation for domain-specific tasks~\citep{Zhao_2019_CVPR,ZHANG2023120041}.
For example, ~\cite{Zhao_2019_CVPR} explores learned data augmentation for biomedical segmentation tasks via labeling of synthesized samples with a single brain atlas. ~\cite{ZHANG2023120041} explores data augmentation in specifically brain segmentation via combining multiple brain scan samples, similarly to Augmix and Cutmix. 
However, this work is reliant on additional annotations to augment regions of interest. 
In our work, we present a generalizable augmentation technique and show that performance boosts generalize well out-of-the-box on several domains.

\begin{table}[t!]
    \centering
    \begin{tabular}{crrr}
        \toprule
        Method & SA Time & Data Gen Time & Storage \\
        \midrule
        AdvSteer & 90.0$\pm$15.5 min & $\sim$ 48 hours & 2.4 TB\\
        Ours & 9.6$\pm$0.2 min & - & 12 GB\\
        \bottomrule
    \end{tabular}
    \vspace*{-0.5em}
    \caption{\textbf{Runtime and Storage Comparison on Sensitivity Analysis of \texttt{AdvSteer}~\citep{yu_shen} vs. Ours.} Our approach enables the practical use of sensitivity analysis in online training as an augmentation policy. We compute each mean and standard deviation value in ``SA Time" across 4 runs. Each sensitivity analysis iteration computes curves for 24 different augmentations at 5 levels each, for a total of 120 evaluation passes. \texttt{Advsteer} requires an offline data generation stage for each dataset, whilst ours is entirely online. Computed SA Time does not factor in data generation time. Ours runs about ${\bf 9.3} \times$ faster and takes ${\bf 200} \times$ less storage in isolation.}
    \label{tab:sa_complexity}
    \vspace*{-1em}
\end{table}

\section{Methodology}
In general, sensitivity analysis examines how small fluctuations in the inputs affects the outputs of a system. 
In our augmentation approach, \textit{the key idea is that sensitivity analysis can be used to sample augmentations uniformly with respect to ``impact on model performance'', as opposed to sampling uniformly across the ``parameterized augmentation space''}. 

To quantify this for a given deep learning model, we need a metric for model performance and a metric for image degradation which is consistent across augmentation types. 
Choosing a model performance metric is straightforward; any bounded measure of accuracy (MA) where higher values are better suffices.
% As for the image degradation metric, we follow that of previous work ~\cite{yu_shen}, which uses distributional metrics to quantify the degree of change from the original data to the perturbed data. However, instead of using Frechet Inception Distance (FID) as done in ~\cite{yu_shen}, 
As for the image degradation metric, we use Kernel Inception Distance (KID), introduced by ~\cite{binkowski_2018} to reduce bias towards sample size. 
At a high level, we use KID to measure the ``distance" between an original dataset and its perturbed version. 
KID does so by passing both datasets through a generalized Inception model, and computing the square Maximum Mean Discrepancy (MMD) between their respective features. 
The reduced sample size bias of KID allows us to approximate the image degradation metric without iterating through the full validation set. \\

\begingroup
\removelatexerror

\SetKwComment{Comment}{/* }{ */}

\begin{algorithm}[t!]
\SetAlgoLined
\SetKwComment{tcc}{// }{}
\KwData{Training dataset $X_t$, Validation dataset $X_v$, Validation Rate $r_v$, SA Rate $r_{SA}$}
\KwResult{Trained semantic segmentation model}
$N_V \gets 0$\ \tcc*{Number of validation rounds}
$f(\cdot) \gets Identity(\cdot)$ \tcc*{Aug transform}
Initialize network weights $\theta$\;

\For{$i \gets 1 ... max\_iter$ \tcc*{Training loop}}{
    $x_{ti} \gets DataLoader(X_t)$\;
    \If{$p_f$ is initialized}{
        $f \sim p_f$ \tcc*{Sample aug PDF}
    }
    $x_{ti}^{aug} \gets f(x_{ti})$\;

    \If{$i$ \% $r_v == 0$}
    {
        \If{$i$ \% $r_{SA} == 0$ \tcc*{Update SA Curve}}
        {
            levels $\gets$ [] \tcc*{Store all $\alpha$ values}
            metrics $\gets$ [] \tcc*{Store all metrics}
            \For{each augmentation type $f$}{
                $\alpha_f, acc_f \gets$ SensitivityAnalysis(f, $\theta$)\tcc*{Appendix: Algorithm~\ref{alg:adaptive}}
                levels.append($\alpha_f$)\;
                metrics.append($acc_f$)\;
            }
            levels = levels.sort() \tcc*{Sort based on descending metrics}
            $p_f \gets$ BetaBinom(idx($f$), 0.75, 1.0) \tcc*{Categorical PDF by Acc}
        }
        \For{$x_{vi} \gets DataLoader(X_v)$ \tcc*{Validation}}{
            Compute clean validation metrics\;
        }
            
    }
}
\caption{Training with Sensitivity-Informed Augmentation.}

\label{alg:training}
\end{algorithm}
\vspace{-1em}

\endgroup

%%%%%%%%%% Wenjie Edits %%%%%%%%%%%
% Our key idea is to use sensitivity analysis to sample augmentations based on their impact on model performance, rather than sampling uniformly across the parameterized augmentation space. 

By sampling augmentations to which the model is sensitive, we can improve robustness. We define the sensitivity of the model to changes in augmentation intensity as the ratio of the change in model accuracy to the change in KID:
\begin{equation} \label{eq:sensitivity_def}
\text{sensitivity} = \frac{\Delta MA}{\Delta KID} 
% \vspace*{-0.5em}
\end{equation}

Our goal is to identify augmentation intensities that result in high sensitivity—that is, small changes in the augmentation (as measured by KID) lead to large changes in model performance (MA). This indicates that the model is particularly sensitive to those augmentations, and training on them could improve robustness.
To formalize this, we seek to find a set of increasing, nontrivial augmentation intensities $\alpha_1 < \alpha_2 < \ldots < \alpha_L$ that maximize sensitivity. We define the local changes in accuracy and KID between consecutive intensities as:

\begin{small}
\vspace*{-1em}
\begin{align} 
\Delta\widehat{MA}( \alpha_i, \alpha_{i-1}) &= MA( \alpha_{i-1}) - MA( \alpha_{i}) \ \\
\Delta\widehat{KID}( \alpha_i, \alpha_{i-1}) &= \frac{D_{\text{KID}}(x_{\alpha_i} \| x_{\text{clean}}) - D_{\text{KID}}(x_{\alpha_{i-1}} \| x_{\text{clean}})}{D_{\text{KID}}(x_{\alpha_{\text{max}}} \| x_{\text{clean}})} 
\end{align}
\end{small}

Here, $MA(\alpha)$ is the model accuracy at augmentation intensity $\alpha$, and $D_{\text{KID}}(x_{\alpha} \| x_{\text{clean}})$ is the KID between the augmented data at intensity $\alpha$ and the original clean data. The normalization in $\Delta\widehat{KID}$ ensures that KID values are comparable across different augmentation types.

We then formulate an objective function $Q$ to find the set of intensities that maximizes sensitivity while ensuring adequate spacing between them:

\begin{equation} \label{eq:q_function}
\begin{aligned}
Q = \underset{{\alpha_1, \ldots, \alpha_L}}{\arg\max} \underset{2 \leq i \leq L}{\min} &\Delta \widehat{MA}( \alpha_i, \alpha_{i-1}) \\
- &\Delta \widehat{KID}( \alpha_i, \alpha_{i-1}) \\
+ &\lambda (\alpha_i - \alpha_{i-1}) 
\end{aligned}
\end{equation}

In this equation, the term $\Delta \widehat{MA}( \alpha_i, \alpha_{i-1})$ represents the change in model accuracy between intensities $\alpha_{i-1}$ and $\alpha_i$. We subtract $\Delta \widehat{KID}( \alpha_i, \alpha_{i-1})$ to favor intensity intervals where accuracy drops more than the image degradation increases, thus indicating higher sensitivity. Furthermore, the regularization term $\lambda (\alpha_i - \alpha_{i-1})$ (with $\lambda > 0$) encourages spacing between intensities, preventing them from being too close together. In our implementation, $\lambda = 2$.

Our objective seeks to maximize the minimum value of this expression across all intervals, ensuring that even the least favorable interval is optimized.

To simplify the optimization, we introduce a function $g(\alpha)$:

\begin{align} \label{eq:g_function} g(\alpha) = 1 - MA(\alpha) - \frac{D_{\text{KID}}(x_{\alpha} \| x_{\text{clean}})}{D_{\text{KID}}(x_{\alpha_{\text{max}}} \| x_{\text{clean}})} + \lambda \alpha \end{align}

% Using $g(\alpha)$, we can express the objective as seeking intensities where the differences $g(\alpha_i) - g(\alpha_{i-1})$ are equal for all $i$. This leads to evenly spaced values along the function $g$.
The set of $\alpha$ values which fulfills $Q$ has the following property: $g(\alpha_2) - g(\alpha_1) = g(\alpha_3) - g(\alpha_2) = ... = g(\alpha_L) - g(\alpha_{L-1})$; in other words, optimal $\alpha$ values are produced at equal intervals along the function $g$.
Since $g(\alpha)$ is approximately monotonically increasing (as $MA(\alpha)$ decreases and $D_{\text{KID}}(x_{\alpha}, x_{\text{clean}})$ increases with increasing $\alpha$), and its values lie within a known range, we can approximate the solution as:

\begin{align} \label{eq:q_solution} \alpha_i \approx g^{-1}\left( \frac{G_{\text{max}} \cdot i}{L} \right), \quad i = 1, \ldots, L \end{align}

where $G_{\text{max}}$ is the maximum value of $g(\alpha)$ over the range of $\alpha$, and $g^{-1}$ is the inverse function.
Since we choose $\lambda = 2$ in our implementation, $G_{\text{max}} = 2$.

However, since we cannot explicitly compute $g^{-1}$ due to $g(\alpha)$ being unknown in closed form, we iteratively estimate the values of $\alpha_i$ using methods like the Piecewise Cubic Hermite Interpolating Polynomial (PCHIP), which is a spline estimation technique. By sampling a few initial points and fitting an interpolating function, we can estimate the intensities that satisfy our objective.
A proof for equal spacing can be found in Appendix Section~\ref{appendix:proof_equal_spacing}.
We show the pseudocode for sensitivity analysis in Algorithm~\ref{alg:adaptive} of the appendix. Additionally, the iterative process for solving $\alpha$ values is visualized in Appendix Figure~\ref{fig:sa_illustration}. 
Below, we show the full training routine involving Sensitivity Analysis in Algorithm~\ref{alg:training}.

\begin{figure*}[t!]
 \vspace{-0.5em}
    \centering
    \subfigure[IDBH.]{\includegraphics[width=0.4\linewidth]{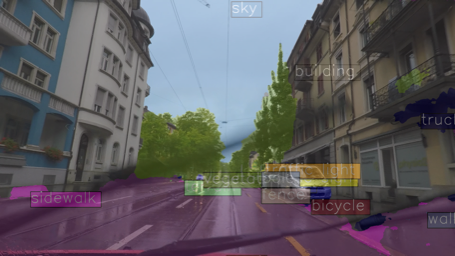}}\quad
    \subfigure[Ours.]
    {\includegraphics[width=0.4\linewidth]{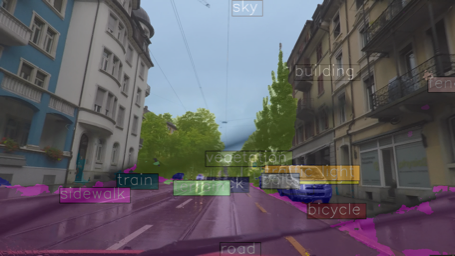}}
\vspace{-1.3em}
    \caption{\textbf{Special case on ACDC prediction: windshield wiper occlusion.} We observe a special case of natural corruptions in rainy weather which cannot be directly simulated by the existing set of perturbations: physical occlusion by windshield wipers. We compare our method to recent SOTA in augmentation for robustness, IDBH~\cite{li2023data}. \textbf{\textit{While IDBH involves random occlusion during training, ours does not. Our augmentation approach achieves comparable qualitative results with a smaller set of augmentations.}} Additional visualizations can be found in Appendix Section~\ref{appendix:qualitative_windshield}.}
    \vspace{-0.5em}
    \label{fig:acdc_fail}
\end{figure*}

\textbf{Resource improvement over previous work in sensitivity analysis.} 
Previous sensitivity analysis methods~\citep{yu_shen} compute $g(\alpha)$ using a uniformly sampled set of $\alpha$ values across the entire augmentation space. This approach requires evaluating the model at many intensities and often necessitates offline generation of augmented datasets for each intensity and augmentation type. As a result, the storage complexity becomes the size of the original dataset multiplied by the number of augmentation types and intensities, leading to substantial storage demands.

In contrast, our method performs sensitivity analysis online during training and adaptively samples intensities based on the model's responses. By estimating $g(\alpha)$ iteratively and focusing only on necessary intensities, we eliminate the need for pre-generating augmented datasets. As a result, our approach only adds about 0.2 * (number of updates) * (evaluation time) amount of time to the total training pipeline, making the use of sensitivity analysis practical for on-the-fly augmentation policy during training.
\begin{table*}[th]
  \centering
  \scalebox{.9}{
  \begin{tabular}{lrrcrrcrrcrr}
    \toprule
      & \multicolumn{2}{c}{Fog} & \phantom{a} & \multicolumn{2}{c}{Rain} & \phantom{a} & \multicolumn{2}{c}{Night} & \phantom{a} & \multicolumn{2}{c}{Snow} \\
   \cmidrule{2-3} \cmidrule{5-6} \cmidrule{8-9} \cmidrule{11-12}
   Method & aAcc$\uparrow$ & mIoU$\uparrow$ & \phantom{a} & aAcc$\uparrow$ & mIoU$\uparrow$ & \phantom{a} & aAcc$\uparrow$ & mIoU$\uparrow$ & \phantom{a} & aAcc$\uparrow$ & mIoU$\uparrow$ \\
   \midrule
   Baseline & 89.70 & 55.10 & \phantom{a} & 87.41  & 42.82 & \phantom{a} & 54.39 & 14.89 & \phantom{a} & 83.23 & 41.22 \\ % updated
   AugMix & 89.76 & 57.79 & \phantom{a} & \textbf{89.28} & 47.53 & \phantom{a} & 56.64 & 17.35 & \phantom{a} & 83.34 & 43.94 \\ % updated
   AutoAugment & 77.06 & 56.18 & \phantom{a} & 75.52 & 42.66 & \phantom{a} & 57.14 & 20.65  & \phantom{a} & 71.83 & 40.94 \\ % updated
   RandAug & 88.24 & 53.99 & \phantom{a} & 86.92 & 43.10 & \phantom{a} & 56.03 & 18.08 & \phantom{a} & 83.35 & 41.86 \\ % updated
   TrivialAug & 85.79 & 55.16 & \phantom{a} & 84.35 & 41.26 & \phantom{a} & 54.52 & 17.02 & \phantom{a} & 77.99 & 42.64 \\
   VIPAug & \textbf{92.04} & 60.28 & \phantom{a} & 89.19 & 46.41 & \phantom{a} & \textbf{61.10} & 17.40 & \phantom{a} & \textbf{85.72} & 45.04 \\
   IDBH & 89.79 & 60.79 & \phantom{a} & 86.93 & 45.64    & \phantom & 54.76 & 18.41 & \phantom{a} & 83.88 & 45.35 \\
   \rowcolor{Gray} Ours & 90.20 & \textbf{62.50} & \phantom{a} & 88.87 & \textbf{49.36} & \phantom{a} & 58.85 & \textbf{20.72} & \phantom{a} & 83.39 & \textbf{48.16} \\
   % \rowcolor{Gray} Ours & \textbf{92.91} & \textbf{67.51} & \phantom{a} & \textbf{89.71} & \textbf{50.97} & \phantom{a} & \textbf{56.32} & 18.07 & \phantom{a} & \textbf{88.47} & \textbf{51.67} \\
   % /fs/nexus-projects/robustness_datasets/iclr_rebuttal/segformer_dataset_experiments_eval
    \bottomrule
  \end{tabular}}
  \caption{
    \textbf{Evaluation of zero-shot adverse weather performance across data augmentation techniques.} We evaluate each data augmentation method across four different weather scenarios from the Adverse Conditions Dataset with Correspondences (ACDC)~\citep{acdc} dataset. Each model is trained only with clean Cityscapes data with the Segformer~\citep{segformer} backbone.
    Our method, highlighted in grey, maintains the best performance across nearly all metrics for three out of four scenarios, with relative mIoU improvement over the next best method of up to \textit{2.81\% on fog, 3.85\% on rain, and 6.20\% on snow}. 
    }
    \vspace{-1em}
    % For Fog: (67.51 - 59.17) / 59.17 * 100 = 14.09
    % Rain: (50.97 - 49.23) / 49.23 * 100 = 3.53
    % Snow: (51.67 - 43.98) / 43.98 * 100 = 17.49
  % \vspace*{-1.5em}
  \label{tb:weather_evaluation}
\end{table*}

\section{Experiments}
\textbf{Hardware.} 
Each experiment is conducted on four NVIDIA RTX A4000 GPUs and 16 AMD Epyc 16-core processors. Sensitivity analysis experiments are conducted on one GPU and 4 processors. 

\textbf{Experiment Setup.}
For evaluation on real-world corruptions and data effiency, we train all experiments with the Segformer~\citep{segformer} backbone, a robust and state-of-the-art architecture for segmentation.
Experiments in downstream fine-tuning from foundation model SAM~\citep{kirillov2023segment}, are restricted to their original ViT~\citep{dosovitskiy2021an_vit} architectures as the backbone.
All experiments, with the exception of downstream fine-tuning experiments, are trained for 160k iterations regardless of approach, and only the best-performing checkpoints by mIoU (mean Intersection-over-Union by class) are used for evaluation in results. 
Fine-tuning experiments initialized from foundation weights are trained for 80k iterations.
Experiments within each table are run with the same hyperparameters with respect to learning configuration, initialization, and architecture.
Additionally, nearly all models share the same set of augmentations, with the exception of IDBH~\citep{li2023data}, which uses an additional two augmentations (RandomFlip and RandomErase). 
We include these two additional augmentations in IDBH experiments to stay faithful to original release implementations by authors.
We use official implementations for each  method, and fix the random seed for each experiment such that they are reproducible. 
More hyperparameter details for experiments can be found in Appendix Section~\ref{appendix:hyperparameter}.
Full experiment configurations will be released alongside the code implementation for full reproducibility of results.

\textbf{Metrics.}
We use three different metrics for evaluating the performance of a segmentation model: absolute pixel accuracy (aAcc), mean pixel accuracy (mAcc), and mean Intersection-over-Union (mIoU). Mean values are taken over object classes---thus, aAcc will be more susceptible to class imbalances, although it is the most intuitive.

\begin{table*}[t!]
  \centering
  \scalebox{.66}{
  \begin{tabular}{llccrrcrrcrrcrr}
    \toprule
      & & & \phantom{a} & \multicolumn{2}{c}{Clean} & \phantom{a} & \multicolumn{2}{c}{Basis Aug} & \phantom{a} & \multicolumn{2}{c}{AdvSteer} & \phantom{a} & \multicolumn{2}{c}{IN-C} \\
   \cmidrule{5-6} \cmidrule{8-9} \cmidrule{11-12} \cmidrule{14-15}
   Dataset & Type & Method & \phantom{a} & aAcc$\uparrow$ & mIoU$\uparrow$ & \phantom{a} & aAcc$\uparrow$ & mIoU$\uparrow$ & \phantom{a} & aAcc$\uparrow$ & mIoU$\uparrow$ & \phantom{a} & aAcc$\uparrow$ & mIoU$\uparrow$ \\
    \midrule
       \multirow{3}{*}{ADE20K} & \multirow{3}{*}{General} & TrivialAug & \phantom{a} & 75.420  &  32.580  & \phantom{ab} &  69.559  &  27.083  & \phantom{ab} &  41.783  &  9.188  & \phantom{ab} &  61.495  &  18.668 \\
       & & IDBH & \phantom{ab} & \textbf{76.220}  &  \textbf{33.950}  & \phantom{ab} &  72.752  &  30.651  & \phantom{ab} &  40.557  &  9.475  & \phantom{ab} &  \textbf{61.971}  &  \textbf{19.091} \\
       & & \cellcolor{Gray} Ours & \cellcolor{Gray} \phantom{a} & \cellcolor{Gray} 76.110  &  \cellcolor{Gray} 33.790  & \cellcolor{Gray} \phantom{ab} &  \cellcolor{Gray} \textbf{74.285}  &  \cellcolor{Gray} \textbf{31.922}  & \cellcolor{Gray} \phantom{ab} &  \cellcolor{Gray} \textbf{43.075}  &  \cellcolor{Gray} \textbf{9.628}  & \cellcolor{Gray} \phantom{ab} &  \cellcolor{Gray} 61.280  &  \cellcolor{Gray} 18.721 \\
       \midrule
       \multirow{3}{*}{VOC2012} & \multirow{3}{*}{General} & TrivialAug & \phantom{a} & 90.090  &  57.900  & \phantom{ab} &  87.837  &  52.340  & \phantom{ab} &  75.350  &  20.338  & \phantom{ab} &  \textbf{82.884}  &  36.080 \\
       & & IDBH & \phantom{ab} & 90.610  &  60.570  & \phantom{ab} &  89.262  &  56.876  & \phantom{ab} &  69.843  &  20.810  & \phantom{ab} &  81.819  &  36.933 \\
       & & \cellcolor{Gray} Ours & \cellcolor{Gray} \phantom{a} & \cellcolor{Gray} \textbf{90.800}  &  \cellcolor{Gray} \textbf{61.140}  & \cellcolor{Gray} \phantom{ab} & \cellcolor{Gray} \textbf{89.555}  & \cellcolor{Gray} \textbf{58.183}  & \cellcolor{Gray} \phantom{ab} & \cellcolor{Gray} \textbf{69.690}  & \cellcolor{Gray} \textbf{21.470}  & \cellcolor{Gray} \phantom{ab} & \cellcolor{Gray} 82.519  & \cellcolor{Gray} \textbf{38.834} \\
       \midrule
       \multirow{3}{*}{POTSDAM} & \multirow{3}{*}{Aerial} & TrivialAug & \phantom{a} & 84.360  &  67.820  & \phantom{ab} &  77.649  &  55.763  & \phantom{ab} &  \textbf{55.817}  &  \textbf{34.282}  & \phantom{ab} &  \textbf{55.866}  &  \textbf{36.967} \\
       & & IDBH & \phantom{ab} & 84.280  &  \textbf{68.690}  & \phantom{ab} &  79.392  &  63.757  & \phantom{ab} &  22.675  &  14.975  & \phantom{ab} &  46.413  &  30.123 \\
       & & \cellcolor{Gray} Ours & \cellcolor{Gray} \phantom{a} & 
       \cellcolor{Gray} \textbf{84.550}  & \cellcolor{Gray} 68.450  & \cellcolor{Gray} \phantom{ab} & \cellcolor{Gray} \textbf{82.590}  & \cellcolor{Gray} \textbf{66.065}  & \cellcolor{Gray} \phantom{ab} & \cellcolor{Gray}  44.817  & \cellcolor{Gray} 29.983  & \cellcolor{Gray} \phantom{ab} & \cellcolor{Gray} 54.275  & \cellcolor{Gray} 36.416 \\
       \midrule
       % \multirow{3}{*}{LoveDA} & \multirow{3}{*}{Aerial} & TrivialAug & \phantom{a} & 70.000  &  50.920  & \phantom{ab} &  66.279  &  44.784  & \phantom{ab} &  38.510  &  19.175  & \phantom{ab} &  48.486  &  27.037 \\
       % & & IDBH & \phantom{ab} & 68.680  &  50.280  & \phantom{ab} &  66.132  &  47.656  & \phantom{ab} &  29.002  &  13.947  & \phantom{ab} &  42.237  &  25.859 \\
       % & & \cellcolor{Gray} Ours & \cellcolor{Gray} \phantom{a} & 68.700  &  48.040  & \phantom{ab} &  66.958  &  46.045  & \phantom{ab} &  38.107  &  16.635  & \phantom{ab} &  48.593  &  27.423 \\
       % \midrule
       \multirow{3}{*}{A2I2Haze} & \multirow{3}{*}{UGV} & TrivialAug & \phantom{a} & 98.730  &  69.180  & \phantom{ab} &  97.317  &  51.800  & \phantom{ab} &  85.598  &  \textbf{22.225}  & \phantom{ab} &  97.363  &  46.502 \\
       & & IDBH & \phantom{ab} & 98.680  &  69.300  & \phantom{ab} &  98.346  &  64.615  & \phantom{ab} &  85.545  &  19.490  & \phantom{ab} &  97.368  &  45.970 \\
        & & \cellcolor{Gray} Ours & \cellcolor{Gray} \phantom{a} & \cellcolor{Gray} \textbf{98.790}  & \cellcolor{Gray} \textbf{70.290}  & \cellcolor{Gray} \phantom{ab} & \cellcolor{Gray} \textbf{98.613}  &  \cellcolor{Gray} \textbf{67.919}  & \cellcolor{Gray} \phantom{ab} & \cellcolor{Gray} \textbf{89.482}  &  \cellcolor{Gray} 21.843  & \cellcolor{Gray} \phantom{ab} & \cellcolor{Gray} \textbf{97.407}  &  \cellcolor{Gray} \textbf{49.805} \\
        \midrule
       \multirow{3}{*}{Cityscapes} & \multirow{3}{*}{Driving} & TrivialAug & \phantom{a} & 95.570  &  74.300  & \phantom{ab} &  86.117  &  56.952  & \phantom{ab} &  69.785  &  \textbf{30.593}  & \phantom{ab} &  82.664  &  44.332 \\
       & & IDBH & \phantom{ab} & 95.530  &  73.930  & \phantom{ab} &  93.160  &  68.052  & \phantom{ab} &  \textbf{71.932}  &  29.388  & \phantom{ab} &  \textbf{83.041}  &  44.225 \\
       & & \cellcolor{Gray} Ours & \cellcolor{Gray} \phantom{a} & \cellcolor{Gray} \textbf{95.780}  & \cellcolor{Gray} \textbf{75.530}  & \cellcolor{Gray} \phantom{ab} & \cellcolor{Gray} \textbf{94.305}  & \cellcolor{Gray} \textbf{71.539}  & \cellcolor{Gray} \phantom{ab} & \cellcolor{Gray} 68.468  & \cellcolor{Gray} 28.070  & \cellcolor{Gray} \phantom{ab} & \cellcolor{Gray} 82.435  & \cellcolor{Gray} \textbf{45.066} \\
       \midrule 
       \multirow{3}{*}{Synapse} & \multirow{3}{*}{Medical} & TrivialAug & \phantom{a} & 98.890  &  62.000  & \phantom{ab} &  97.939  &  49.237  & \phantom{ab} &  \textbf{97.243}  &  \textbf{32.182}  & \phantom{ab} &  98.425  &  51.512 \\
       & & IDBH & \phantom{ab} & 99.150  &  67.720  & \phantom{ab} &  98.912  &  63.504  & \phantom{ab} &  95.143  &  29.760  & \phantom{ab} &  \textbf{98.486}  &  53.475 \\
       & & \cellcolor{Gray} Ours & \cellcolor{Gray} \phantom{a} & \cellcolor{Gray} \textbf{99.250}  & \cellcolor{Gray} \textbf{71.380}  & \cellcolor{Gray} \phantom{ab} & \cellcolor{Gray} \textbf{99.082}  & \cellcolor{Gray} \textbf{68.828}  & \cellcolor{Gray} \phantom{ab} & \cellcolor{Gray} 90.282  & \cellcolor{Gray} 30.310  & \cellcolor{Gray} \phantom{ab} & \cellcolor{Gray} 96.779  & \cellcolor{Gray} \textbf{56.013} \\
    \bottomrule
  \end{tabular}}
  % \vspace*{-1.5em}
 \caption{
\textbf{Performance evaluation of our method vs. SOTA on synthetic scenarios across 6 different datasets.} We evaluate our method and SOTA on ADE20K~\citep{zhou2019semantic_ade20k}, VOC2012~\citep{voc2012}, POTSDAM~\citep{potsdam}, A2I2Haze~\citep{a2i2haze}, Cityscapes~\citep{Cordts2016Cityscapes}, and Synapse~\citep{synapse} datasets, across three synthetic corruption scenarios: individual basis augmentations (Basis Aug), compositions of photometric augmentations produced by sensitivity analysis in Adversarial Steering (AdvSteer) ~\citep{yu_shen}, and the synthetic augmentation benchmark ImageNet-C (IN-C)~\citep{imagenet-c}. 
{\em Our method consistently achieves improved performance on synthetic corruption benchmarks while still maintaining or even improving clean evaluation accuracy}.
}
\vspace{-1em}
  \label{tb:dataset_experiment_comparison_small}
\end{table*}

\subsection{Evaluation on Real-World Corruptions}
To evaluate the robustness of our model in visual and graphics applications, we test on real-world adverse samples. 
While real-world adverse samples in most datasets are difficult to obtain, there are numerous real-world datasets for driving representing different cities and adverse weather scenarios. 

% We evaluate Cityscapes models with Segformer backbone on two real-world datasets: the Adverse Conditions Dataset with Correspondences (ACDC)~\citep{acdc} dataset which represents adverse weather, and the India Driving Dataset (IDD)~\citep{idd} which represents an alternative, more heterogeneous domain. 
% IDD represents an alternative, but similar, domain in which visual appearances of vehicles, traffic, and scenery may slightly change, in addition to co-occurrences of classes. 

We compare our results to seven methods: a baseline model where no augmentation is performed, AugMix~\citep{hendrycks_2020}, AutoAugment~\citep{autoaugment}, RandAugment~\citep{cubuk_2020}, and TrivialAugment~\citep{muller_2021}, VIPAug~\cite{lee2024_vipaug}, and IDBH~\citep{li2023data}.
The Inception model used to compute KID in our method is pre-trained on ImageNet; likewise, for policy-based augmentation techniques such as Augmix and AutoAugment, the policies are also based off of ImageNet.
Different from other policy-based methods, our approach estimates the current model's sensitivity to perturbations relative to the Inception model pre-trained on ImageNet and utilizes the information in augmentation sampling; no additional ``policy" parameters are trained.
On real-world dataset evaluation for unseen weather and domain gap scenarios, our method shows improvements over the next best performing model across almost all metrics.
We include a qualitative visualization of our model versus several other methods in Figure~\ref{fig:acdc_rainy} of the appendix, which shows inference on a rainy weather sample.
Amongst all methods, a common failure mode is the presence of windshield wipers in rainy weather. 
A visualization of this can be found in Appendix Section~\ref{appendix:failure_acdc}.

A break-down the performance on the ACDC dataset by weather type in Table~\ref{tb:weather_evaluation}. 
In total, the ACDC dataset has four different weather scenarios: Fog, Rain, Night, and Snow, where the largest relative boost over next-best method, IDBH~\cite{li2023data}, (6.20\%) is in Snow scenarios.
In three out of four weather categories, our method outperforms other methods, with the exception of Night scenarios.
AugMix achieves higher aAcc but lower mIoU than our method on Rain scenarios possibly due to class imbalances, such as the large number of pixels classified as ``sky". While the total \# of correct pixels is higher on AugMix, our method outperforms when averaged by class, on mIoU.
Night scenario visibility corruption stems from lack of lighting, as opposed to the other three, which may have more differences in object appearances and blurring effects.
While our method does not perform worse in mIOU, we do perform worse in aACC. 
This may suggest that the failure mode of our method in Night scenarios are due to smaller objects covering less pixel space. 
Performance on both ACDC and the India Driving (IDD)~\cite{idd} datasets across multiple methods can be found in Section~\ref{appendix:realworld} of the Appendix. 

\textbf{Special case: co-occurence of windshield wipers and rainy weather.} In the ACDC dataset, the rainy scenario evaluation set contains co-occurences with windshield-wiper occlusion. 
This case is interesting in that occlusions are not included in any experiments except those of IDBH. In qualitative results, we observe that our method handles windshield wiper occlusions just as well, if not better, than IDBH. In Figure~\ref{fig:acdc_fail}, we show an example of this, where our method shows comparatively less artifacts in the building and sky, despite not having been trained on occlusion (RandomErase) augmentations.

\begin{figure*}[t!]
  \centering
  \subfigure[\# Samples vs. ACDC mIoU performance.]{\includegraphics[width=0.45\linewidth]{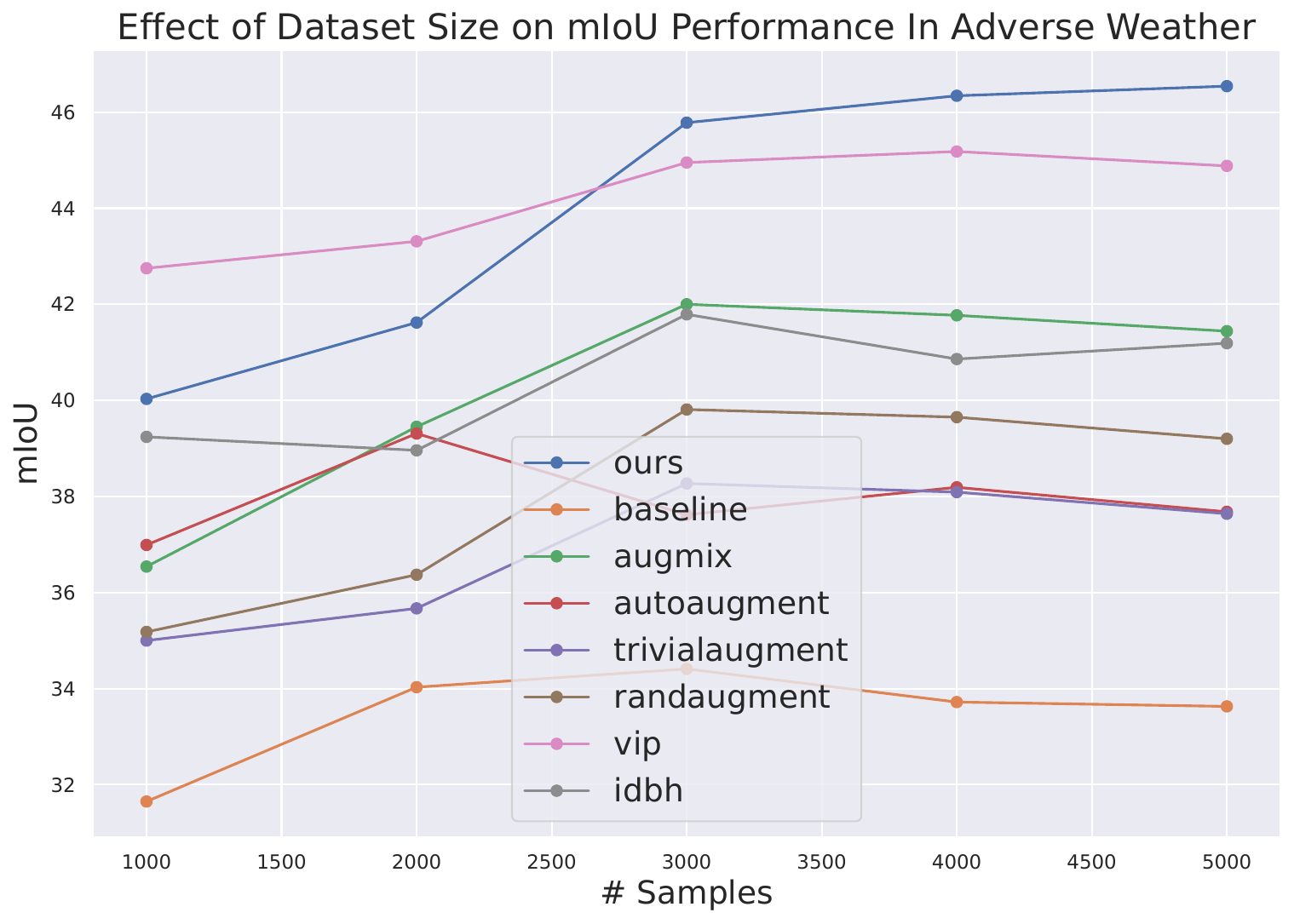}}\quad
  \subfigure[\# Samples vs. IDD mIoU performance.]{\includegraphics[width=0.45\linewidth]{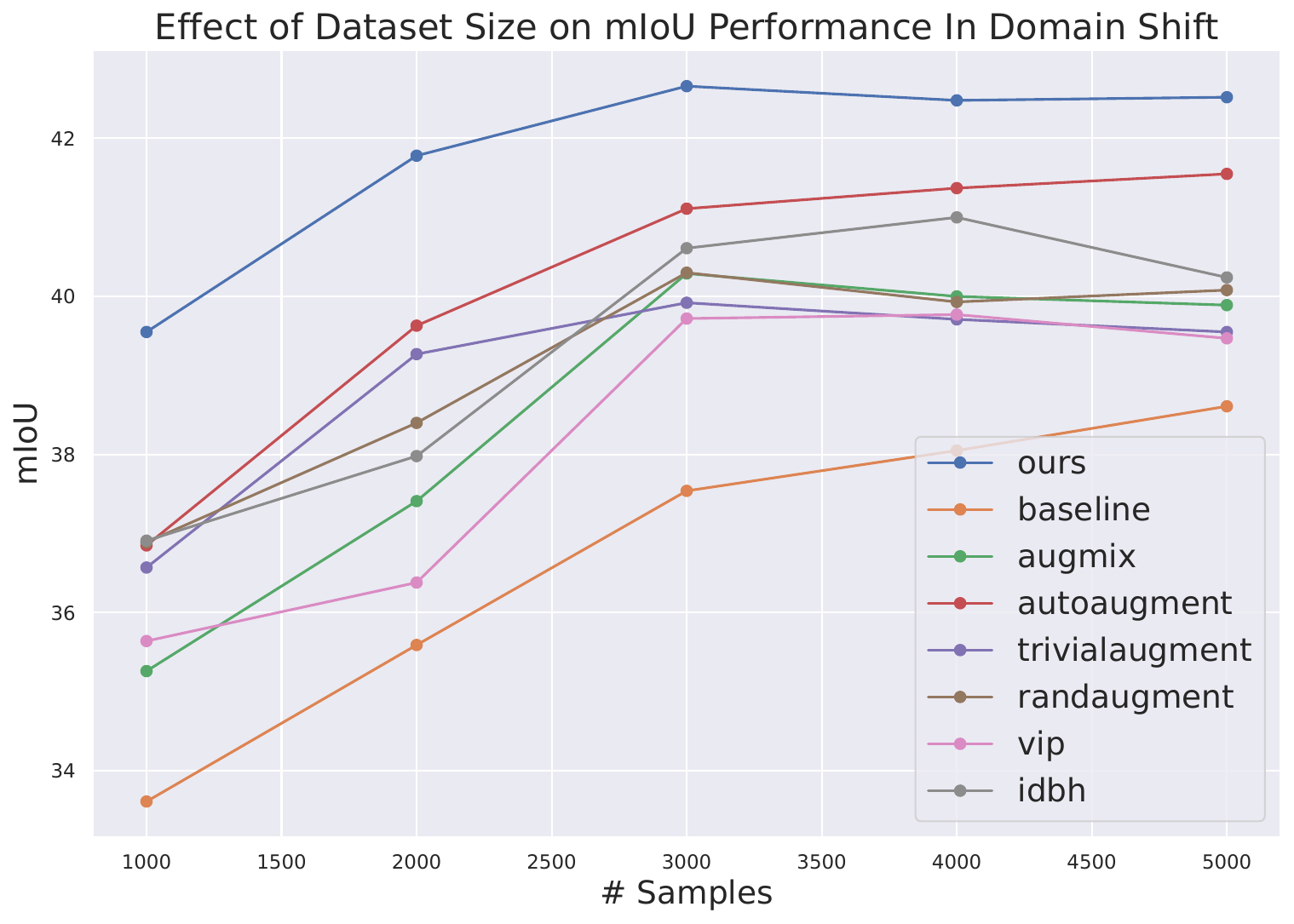}}
  \vspace*{-1em}
  \caption{{\bf Comparison of Ours vs. Other SOTA Methods}:  Ours (top, \textcolor{blue}{blue}) outperforms all others in performance as the number of samples increases, while other methods plateau on both (a) adverse weather data (ACDC) and domain shift (IDD).}
  \vspace{-1em}
  \label{fig:data_scaling}
\end{figure*}

\subsection{Evaluation on Datasets}
The results of previous experiments show the efficacy of our method in context of driving domains. 
In this experiment, we demonstrate that our method also shows improvements across several datasets and visual computing domains compared to SOTA. 

We evaluated our method on six semantic segmentation datasets, ranging from generic everyday scenes to application-specific domains like driving or medicine. ADE20K~\citep{zhou2019semantic_ade20k}, VOC2012~\citep{voc2012}, POTSDAM~\citep{potsdam}, 
% LoveDA~\citep{loveda}, 
Cityscapes~\citep{Cordts2016Cityscapes}, Synapse~\citep{synapse}, and A2I2Haze~\citep{a2i2haze}. 
% POTSDAM is a remote sensing datasets taken from aerial views, with classes focusing on classification of buildings, roads, trees, etc. POTSDAM describes aerial imagery in Potsdam, Germany. Cityscapes is a popular benchmark dataset for segmentation in urban traffic scenes, with annotations describing classes such as terrain, human, and vehicle types. ADE20K and VOC2012 are generic datasets describing everyday life and objects, with both indoor and outdoor scenes. Synapse is a medical imaging dataset of clinically-acquired CT scans. In our experiments, we use abdomen data and classify organs. A2I2Haze is a dataset representing outdoor clear and hazy data collected from unmanned robots for scene understanding. We use the UGV, or Unmanned Ground Vehicle data in our experiments, which is similar to autonomous driving datasets except in more heterogeneous outdoor environments.
In Table~\ref{tb:dataset_experiment_comparison_small}, we show mIoU performance of our method versus the next-best augmentation technique, the SOTA baseline. 
We evaluate on clean data and three different synthetic scenarios: individual transformations from the basis augmentations at uniform parameter intervals (Basis Aug), the combined perturbation benchmark from ~\cite{yu_shen} (AdvSteer), and ImageNet-C (IN-C)~\citep{imagenet-c} corruptions. 
\textit{On the synthetic benchmark ImageNet-C~\citep{imagenet-c}, our model achieves improved scores, particularly in the robotics and medical domains.}
Our method performed worse primarily in the AdvSteer benchmark of Table~\ref{tb:dataset_experiment_comparison_small}, notably for Cityscapes and Synapse.
This may be due to the sheer intensity of benchmark corruption---the AdvSteer benchmark applies a combination of intense perturbations (not the same as the augmentations used during training), resulting in an extreme case from the original distribution. 
This may be related to degraded performance on Night scenarios in ACDC evaluation, as both scenarios heavily corrupt visibility based on color. 
Examples of the AdvSteer benchmark corruptions can be found in Appendix Section~\ref{appendix:advsteer_examples}.
% Qualitative results on Synapse with synthetic motion blur between our method and next best, TrivialAugment, can be observed in Figure~\ref{fig:synapse_qualitative} of the appendix. 
We emphasize that our method is not necessarily bound to image segmentation---we find similar boosts in performance in preliminary experiments with classification (see Appendix Section~\ref{tab:cub_classification_experiment}).

\begin{table}[t!]
    \centering
    \begin{tabular}{lrrr}
        \toprule
        \phantom{a} & \multicolumn{3}{c}{ViT+SAM} \\
        \cmidrule{2-4}
        Method & aAcc$\uparrow$ & mAcc$\uparrow$ & mIoU$\uparrow$ \\
        \midrule
        Baseline & 84.93 & 62.84 & 52.20 \\
        AugMix & 84.69 & 63.25 & 54.18\\
        AutoAugment & 85.17 & 61.28 & 53.11 \\
        RandAug & 85.16 & 59.33 & 51.95 \\
        TrivialAugment & 84.87 & 59.92 & 50.58 \\
        VIPAug & 84.29 & 61.00 & 51.93 \\
        IDBH & 85.14 & 62.82 & 54.35 \\
        \rowcolor{Gray} Ours & \textbf{85.37} & \textbf{65.18} & \textbf{54.84} \\
        \bottomrule
    \end{tabular}
    \caption{\textbf{Performance on ACDC when fine-tuning downstream segmentation with SAM.} We show additional comparisons when initialized with SAM weights, similarly to results in Table~\ref{tab:foundation_finetuning}.}
    \label{tab:sam_ACDC_finetune}
    \vspace*{-1em}
\end{table}

\begin{table}[t!]
    \centering
    \begin{tabular}{lrrr}
        \toprule
        Method & aAcc$\uparrow$ & mIoU$\uparrow$ & mAcc$\uparrow$ \\
        \midrule
        Baseline & 94.19 & 66.67 & 75.28 \\
        AugMix & 93.99 & 66.07 & 73.60\\
        AutoAugment & 93.84 & 64.99 & 72.09 \\
        RandAug & 92.60 & 59.37 & 66.03 \\
        TrivialAugment & 93.55 & 65.03 & 71.71\\
        VIPAug & 92.98 & 63.67 & 70.86 \\
        IDBH & 93.62 & 65.49 & 74.96 \\
        \rowcolor{Gray} Ours & \textbf{93.88} & \textbf{68.03} & \textbf{75.96} \\
        \bottomrule
    \end{tabular}
    \caption{\textbf{Domain adaptation; fine-tuning on a small target domain dataset.} We show performance on the validation set of the ACDC Snow dataset, after training for 20k iterations on the ACDC Snow training set. All experiments are initialized with a Segformer-b0 model pre-trained on Cityscapes.}
    \label{tab:domain_adaptation}
    \vspace*{-1em}
\end{table}

\subsection{Downstream Finetuning with Foundation Models}
A popular choice for boosting feature robustness is fine-tuning downstream tasks from foundation models.
In these experiments, we examine how our approach can complement robustness provided by foundation models when fine-tuning on downstream tasks. 
We first initialize a distilled SAM~\citep{kirillov2023segment} model on the ViT-Small (ViT-S) architecture, then fine-tune on the semantic segmentation task with Cityscapes.
We choose Cityscapes due to the availability of real-world corrupted images (ACDC and IDD) to evaluate on. 
In our experiments, we observe the highest performance on our method all three metrics.
While the largest boost in robustness stem from robust foundation model features, our results suggest that our method can complement approaches centered around model architecture (such as Segformer). 
Additional results on downstream finetuning with DinoV2 can be found in Table~\ref{tab:foundation_finetuning} of the Appendix.

\begin{figure*}[t!]
    \centering
    % \vspace*{-1em}
    \includegraphics[width=0.9\linewidth]{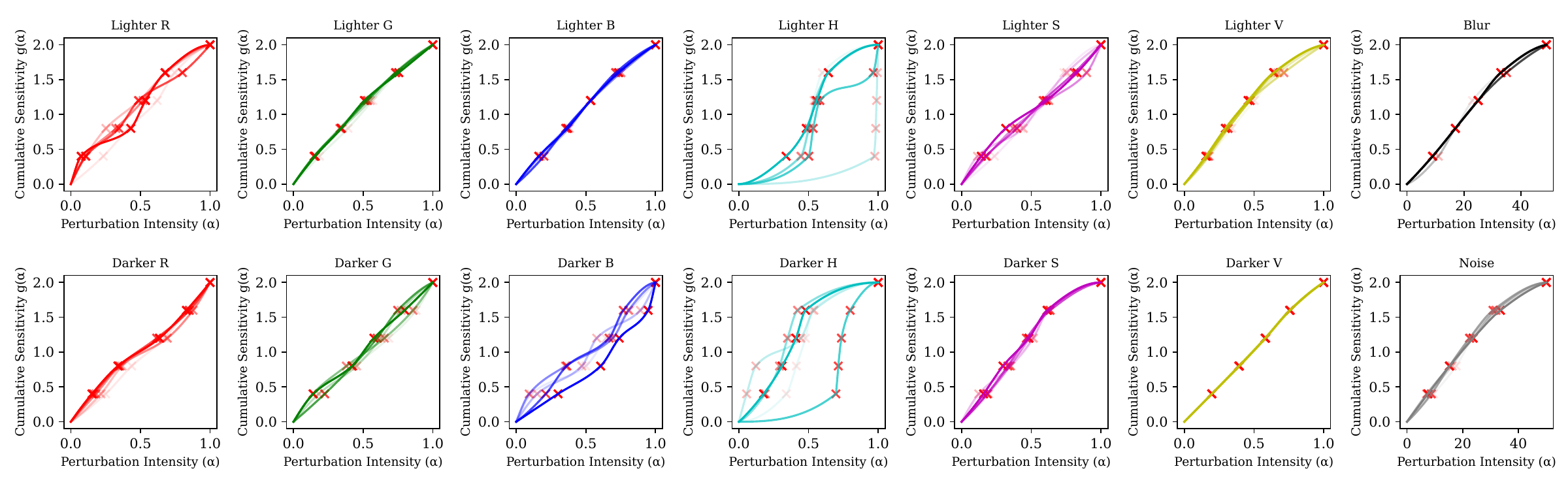}
    \vspace*{-1.5em}
    \caption{\textbf{Sensitivity curves (g values) for color channels, noise, and blur throughout training of Cityscapes.} We visualize how the estimated cumulative sensitivity curve, Equation~\ref{eq:g_function}, changes for RGB, HSV, Gaussian blur, and Gaussian noise during augmented training. In this plot, the most recent curve is opaque, while others decrease in opacity in order of recency. The red X markers indicate the values at which $\alpha$ values are selected (horizontal axes). Surprisingly, most curves remain largely stagnant throughout training, with the exception of \textcolor{teal}{Hue} in HSV (\textcolor{teal}{teal}, center), which changes drastically as the model generalizes. This may suggest that {\em Hue is a highly adversarial factor in model generalization}. Ablation study results in Table~\ref{tb:ablation_study2_small} support this, where the model trained without photometric augmentations demonstrate a significant decrease in performance.}
    \label{fig:sa_curve_training}
    % \vspace*{-1em}
\end{figure*}

\begin{table*}[t!]
  \centering
  \scalebox{.8}{
  \begin{tabular}{lcrrcrrcrrcrrcrr}
    \toprule
      & \phantom{a} & \multicolumn{2}{c}{Clean} & \phantom{a} & \multicolumn{2}{c}{Basis Aug} & \phantom{a} & \multicolumn{2}{c} {AdvSteer} & \phantom{a} & \multicolumn{2}{c}{IN-C} & \phantom{a} & \multicolumn{2}{c}{ACDC}\\
   \cmidrule{3-4} \cmidrule{6-7} \cmidrule{9-10} \cmidrule{12-13} \cmidrule{15-16}
   Method & \phantom{a} & aAcc$\uparrow$ & mIoU$\uparrow$ & \phantom{a} & aAcc$\uparrow$ & mIoU$\uparrow$ & \phantom{a} & aAcc$\uparrow$ & mIoU$\uparrow$ & \phantom{a} & aAcc$\uparrow$ & mIoU$\uparrow$ & \phantom{a} & aAcc$\uparrow$ & mIoU$\uparrow$  \\
    \midrule
        Baseline & \phantom{a} & 95.610  &  75.130  & \phantom{ab} &  92.042  &  65.319  & \phantom{ab} &  62.040  &  21.995  & \phantom{ab} &  79.437  &  38.362 & \phantom{a} & 78.49 & 37.54 \\
        Ours$_{\sim g}$ & \phantom{a} & 95.780  &  \textbf{75.500}  & \phantom{ab} &  93.405  &    68.877  & \phantom{ab} &  \textbf{71.070}   &  27.997  & \phantom{ab} &  83.032  &    44.385 & \phantom{ab} & 78.13 & 43.69 \\
        Ours$_{\sim p}$ & \phantom{a} & 95.740  &  75.210  & \phantom{ab} &  92.544  &  69.002  & \phantom{ab} &  64.907   &  22.437  & \phantom{ab} &  80.817  &    40.876 & \phantom{ab} & 75.74 & 37.97 \\
        Ours$_{\sim Warmup}$ & \phantom{a} & \textbf{95.830}  &  75.430  & \phantom{ab} &  \textbf{94.458}  &    \textbf{71.891}  & \phantom{ab} &  69.138  &  28.472  & \phantom{ab} &  84.438  &    45.849 & \phantom{ab} & 79.78 & 44.66 \\
        Ours$_{\sim Uniform}$ \phantom{a} & & 95.740  &    75.200  & \phantom{ab} &  94.304  &  71.213  & \phantom{ab} &  69.678  &    27.235  & \phantom{ab} &  \textbf{85.135}  &  \textbf{46.219} & \phantom{ab} & \textbf{80.95} & 43.17 \\
        \rowcolor{Gray} Ours & \phantom{a} & 95.790  &  75.100  & \phantom{ab} &  94.439  &    71.665  & \phantom{ab} &  70.605  &  \textbf{28.895}  & \phantom{ab} &  83.844  &    45.617 & \phantom{ab} & 80.13 & \textbf{44.67} \\ % uniform sampling
    \bottomrule
  \end{tabular}}
 \caption{{\textbf{Ablation study results} comparing different variants of our method. We compare: (1) a baseline trained with no augmentations, (2) a variant of our method that only augments with photometric augmentations (Ours$_{\sim g}$), (3) a variant of our method that only uses geometric augmentations (Ours$_{\sim p}$), (4) a variant of our method trained without clean training warm-up, (5) a variant of our method with uniform augmentation (Ours$_{Uniform}$) of computed sensitivity analysis values $\alpha$,  and (6) our full method combining informed probability sampling, and adaptive sensitivity analysis, and all augmentation types (Ours).
}}
\label{tb:ablation_study2_small}
\vspace{-1em}
\end{table*}

\subsection{Domain Adaptation With Small Datasets}
To evaluate the efficacy of our approach in domain adaptation settings, we finetune Segformer Cityscapes models on the training set of the ACDC snow data, then evaluate on the validation set of the ACDC snow data. 
The main difference from previous experiments here is that the model has been trained on target domain data, rather than being evaluated in a zero-shot setting.
These results can be found in Table~\ref{tab:domain_adaptation}.
Interestingly, the baseline approach seems to work second-best for domain adaptation, behind our method. We suspect that the degradation in performance is due to overfitting; with smaller datasets and less challenging augmentations, the model may be more prone to overfitting than generalization. Our method may mitigate this by sampling evenly difficult augmentations with respect to the current model state.

\subsection{Data Efficiency}

We also analyze data efficiency of our method in comparison to other data augmentation methods by training various Segformer models with varying training dataset sizes. 
For each augmentation method, we train five models with training dataset sizes of 1000, 2000, 3000, 4000, and 5000 samples from the Cityscapes dataset. 
We plot the progression of mIoU (Minimum Intersection over Union) performance (higher the better) on 
(a) adverse weather data (ACDC) and (b) the domain shift setting (IDD), as shown in Figure~\ref{fig:data_scaling}.
Our method, in blue, shows consistent improvement on adverse weather and domain shift evaluation with increasing number of samples, and maintains best mIoU performance across each \# of samples slice, suggesting that {\em our method is more data efficient than others}.
Interestingly, not all methods show increased robustness to adverse weather as the number of samples increases for training, indicating that {\em in some cases, scaling data may not necessarily mean increased robustness}.

\subsection{Ablation Study}
We examine several variants of our method to determine the impact of individual components in an ablation study: a baseline trained only with random cropping, a variant of our method using only geometric augmentations, a variant of our method using only photometric augmentations, a variant of our method without clean training warm-up, and a variant of our method using uniform sampling instead of the Beta-Binomial sampling described in Algorithm~\ref{alg:training}.
Uniform sampling of augmentation parameters computed with sensitivity analysis decreases generalization to both synthetic and real-world corruption benchmarks by small margins. 
In addition, training without clean warm-up produces similar results to that with warm-up, suggesting that warm-up is optional.
In our case, warming up with clean evaluation reduces the total number of sensitivity analysis updates, making warm-up with clean evaluation marginally less resource expensive ($\sim$ 0.5 GPU hours total). 
Interestingly, while clean performance remains largely the same across all models, the largest decrease in performance on unseen corruption benchmarks comes from the lack of photometric augmentations. 

\subsection{Color Channel Sensitivity Across Training}
Sensitivity analysis can also be useful for interpretable analysis on-the-fly during training, aside from being used in augmentations directly. 
To analyze how sensitivity changes with respect to different channel values, we plot each $g$ curve computed from Equation~\ref{eq:g_function} for RBG, HSV, Noise, and Blur channels across training for our Cityscapes experiments in Figure~\ref{fig:sa_curve_training}. 
Note that the curves in this figure are based on individual color channels and are separate from those used during training for generalization analysis purposes.
Computed $\alpha$ values will center towards regions with higher sensitivity relative to the current model.
From this visualization, we observe that Hue curves (teal, center) are most volatile during training, with most sensitive augmentation parameters falling towards $\alpha$ values close to 1.0 in the beginning of training. 
As the model generalizes, the Hue curve converges slowly towards $\alpha$ values centered around 0.5, similarly to other curves.
\textit{This suggests that Hue may be a significant adversarial factor in model robustness} compared to other channels.

\section{Discussion and Conclusion}
In this paper, we present a method for sensitivity-informed augmented training for semantic segmentation. Our work is inspired by applications in real-time perception systems such as robotics, where natural corruptions may be abundant at inference time.
Our method combines the information granularity of sensitivity analysis-based methods and the scalability of data augmentation methods, which run on-the-fly during training.
In our results, we show that our method achieves improved robustness on zero-shot real-world adverse weather and domain shift scenarios, in addition to improvements on synthetic benchmarks like ImageNet-C. Additionally, evaluation on real world datasets show clear improvements over current SOTA methods for augmentation. 
Our model can complements other approaches for model robustness such as architecture design and downstream fine-tuning with foundation models.

\noindent
{\bf\large Acknowledgement:}  This research is supported in part by the U.S.
Army Research Labs Cooperative Agreement on ``{\em AI and Autonomy for Multi-Agent Systems}''.

\section*{Impact Statement} 
The goal of this work is to introduce an augmentation framework which enhances generalization to naturally-occurring corruptions.
Our work is driven by problems in real-time systems such as robotics. 
To the best of our knowledge, combating natural corruptions has only positive implications for the systems which it is a concern; for example, autonomous vehicles may be more robust to adverse weather or unexpected scenarios. 

% Currently, our method does not address gaps in low-lighting scenarios. 
% Future work can explore occlusion and low-lighting techniques for segmentation, as both cases resulted in degraded performance for all methods.
% Additionally, our method treats all types of data augmentation as equal, in that the weighting of the augmentation is uniform between types, and sensitivity analysis is used to update the intensity values $\alpha$ only for online sampling. From our ablation study, we show that uniform sampling matters little in context of our method. However, future work dissecting whether all augmentations are equal, especially photometric augmentations, will be useful for unseen scenarios in robotics.

% Future work can further develop this by analyzing the inequality of augmentations dynamically during training.

\bibliography{references}
\bibliographystyle{icml2025}

%%%%%%%%%%%%%%%%%%%%%%%%%%%%%%%%%%%%%%%%%%%%%%%%%%%%%%%%%%%%%%%%%%%%%%%%%%%%%%%
%%%%%%%%%%%%%%%%%%%%%%%%%%%%%%%%%%%%%%%%%%%%%%%%%%%%%%%%%%%%%%%%%%%%%%%%%%%%%%%
% APPENDIX
%%%%%%%%%%%%%%%%%%%%%%%%%%%%%%%%%%%%%%%%%%%%%%%%%%%%%%%%%%%%%%%%%%%%%%%%%%%%%%%
%%%%%%%%%%%%%%%%%%%%%%%%%%%%%%%%%%%%%%%%%%%%%%%%%%%%%%%%%%%%%%%%%%%%%%%%%%%%%%%
\newpage
\appendix

% uncomment when done working
\onecolumn
\section*{Appendix}
\section{Proof on Equal Spacing}
\label{appendix:proof_equal_spacing}

\begin{enumerate}
	\item $\alpha_0 = 0$ represents an augmentation intensity of $0$, i.e. a clean image.
	\item $MA(0)$ approaches $1$ as models get better, but to be more precise, the function $g(\alpha)$ should be
$$g(\alpha) = MA(0) - MA(\alpha) - \frac{D_{\text{KID}}(x_{\alpha} \| x_{\text{clean}})}{D_{\text{KID}}(x_{\alpha_{\text{max}}} \| x_{\text{clean}})} + \lambda \alpha.$$
	\item $Q$ is the set of points $\{\alpha_1, \ldots, \alpha_L\}$ that maximizes the following:
\begin{align*}
g' &= \underset{1 \leq i \leq L}{\min} \widehat{MA}(\alpha_i, \alpha_{i-1}) - \Delta \widehat{KID}(\alpha_i, \alpha_{i-1}) + \lambda (\alpha_i - \alpha_{i-1}) \\
&= \underset{1 \leq i \leq L}{\min} g(\alpha_i) - g(\alpha_{i - 1}).
\end{align*}
\item From the definition of $g(\alpha)$, we note
\begin{align*}
	g(0) &= 0 \\
    g(\alpha_L) &= G_{\text{max}} = (MA(0) - MA(\alpha_L)) - 1 + \lambda \alpha_L.
\end{align*}
In our implementation, we normalize $MA$ so that $g(\alpha_L) = 2$.
\end{enumerate}

\textbf{Proof of equal-spacing:}

We will prove that $g' = G_{\text{max}}/L$, i.e. given the set of $\alpha_i$ values that fulfills $Q$, the set of $g(\alpha_i)$ are equally spaced along the $y$-axis of the function $g(\alpha)$.
BWOC, assume $g' > G_{\text{max}}/L$. Then, $\forall i, g(\alpha_i) - g(\alpha_{i - 1}) \geq g' > G_{\text{max}}/L$.
Taking the sum over $i$ yields a contradiction:
$$G_{\text{max}} = g(\alpha_L) - g(0) = \sum_{i = 1}^L g(\alpha_i) - g(\alpha_{i - 1}) > L\left(\frac{G_{\text{max}}}{L}\right) = G_{\text{max}}.$$
Thus, $g' \leq G_{\text{max}}/L$. If we assume $g(\alpha)$ is continuous over $[0, \alpha_L]$, then by the Intermediate Value Theorem, $\forall i, \exists a_i$ such that
$$g(a_i) = \frac{G_{\text{max}} \cdot i}{L}.$$
Then, $\forall i, g(\alpha_i) - g(\alpha_{i - 1}) = G_{\text{max}}/L$, so the maximum $g' = G_{\text{max}}/L$ can be attained for a specific set of points $Q$.
In the paper, we assume $g(\alpha)$ is strictly monotonically increasing because $MA(\alpha)$ decreases and $D_{\text{KID}}(x_\alpha, x_\text{clean})$ increases as $\alpha$ increases. Based on this, we obtain formula 6.

\section{Sensitivity Analysis Pseudocode}

\begingroup
\removelatexerror

\SetKwComment{Comment}{/* }{ */}

% \begin{tcolorbox}[fonttitle=\bfseries, title=Fast Sensitivity Analysis]

\begin{algorithm}[H]
\SetAlgoLined
\SetKwRepeat{Loop}{loop}{end}
\KwData{Number of levels $L$, Uncertainty threshold $\epsilon$}
\KwResult{Perturbation Levels $\{\alpha_1, ..., \alpha_{L-1}\}$}
$g(\alpha) \gets $ Equation~\ref{eq:g_function}\;
points $\gets \{(0,0), (\alpha_L, 2)\}$\;
\Loop{}{
    $\hat{c} \gets \text{PCHIP}(\text{points})$\;
    \For{$i \gets 1 ... L-1$}{
        $\alpha_i \gets$ Estimate($\hat{c}$, $2i/L$)\;
        $(y_l, y_u) \gets$ Estimate upper and lower y-values of $\hat{c}$ at $x = \alpha_i$\;
        $\hat{c}_l \gets \text{PCHIP}(\text{points.insert}(y_l))$\;
        $\hat{c}_u \gets \text{PCHIP}(\text{points.insert}(y_u))$\;
        $\alpha_{i_l} \gets$ Estimate($\hat{c}$, $y_l$)\;
        $\alpha_{i_u} \gets$ Estimate($\hat{c}$, $y_u$)\;
        $\epsilon_i$ $\gets (\alpha_{i_u} - \alpha_{i_l}) / 2$\;
    }
    $\alpha^*, \epsilon^* \gets$ Choose level with max $\epsilon_i$\;
    \lIf{$\epsilon^* < \epsilon$}{Break loop}
    points.insert$((\alpha^*, g^*(\alpha^*)))$\;
}
\caption{Fast Sensitivity Analysis}
\label{alg:adaptive}
\end{algorithm}

% \begin{algorithm}[H]
% \SetAlgoLined
% \SetKwComment{tcc}{// }{}
% \KwData{Training dataset $X_t$, Validation dataset $X_v$, Validation Rate $r_v$, SA Rate $r_{SA}$}
% \KwResult{Trained semantic segmentation model}
% $N_V \gets 0$\ \tcc*{Number of validation rounds}
% $f(\cdot) \gets Identity(\cdot)$ \tcc*{Augmentation transformation}
% Initialize network weights $\theta$\;

% \For{$i \gets 1 ... max\_iter$ \tcc*{Training loop}}{
%     $x_{ti} \gets DataLoader(X_t)$\;
%     \If{$p_f$ is initialized}{
%         $f \sim p_f$ \tcc*{Sample aug PDF}
%     }
%     $x_{ti}^{aug} \gets f(x_{ti})$\;

%     \If{$i$ \% $r_v == 0$}
%     {
%         \If{$i$ \% $r_{SA} == 0$ \tcc*{Update Sensitivity Analysis}}
%         {
%             levels $\gets$ [] \tcc*{Store all $\alpha$ values}
%             metrics $\gets$ [] \tcc*{Store all metrics}
%             \For{each augmentation type $f$}{
%                 $\alpha_f, acc_f \gets$ SensitivityAnalysis(f, $\theta$)\;
%                 levels.append($\alpha_f$)\;
%                 metrics.append($acc_f$)\;
%             }
%             levels = levels.sort() \tcc*{Sort based on descending metrics}
%             $p_f \gets$ BetaBinom(idx($f$), 0.75, 1.0) \tcc*{Categorical PDF by Acc}
%         }
%         \For{$x_{vi} \gets DataLoader(X_v)$ \tcc*{Validation loop}}{
%             Compute clean validation metrics\;
%         }
            
%     }
% }
% \caption{Training with Sensitivity-Informed Aug}
% \label{alg:adaptive}
% \end{algorithm}

\endgroup

% \begin{algorithm}[tb]
%    \caption{Fast Sensitivity Analysis}
%    \label{alg:adaptive}
% \begin{algorithmic}
%    \STATE {\bfseries Input:}  Number of levels $L$, Uncertainty threshold $\epsilon$
%    \STATE $MA_0$ \leftarrow MA of Clean Dataset
%    \STATE $MA_L$ \leftarrow MA of Dataset Max Perturbation
%    \STATE $KID_L$ \leftarrow KID of Dataset Max Perturbation
%    \STATE Points: [(0,0), (maxlevel, 2)]
%    \REPEAT
%    \STATE Curve = PCHIP(points)
%    \STATE $Q$ = estimate from curve
%    \FOR{level $\in$ $Q$}
%    \STATE Obtain higher and lower y-bounds (use monotonicity of curve)
%    \STATE Interpolate curve using PCHIP with higher and lower guess
%     \STATE Uncertainty = difference between new level estimates with these curves   \ENDFOR
%     \IF{uncertainty is less than threshold}
%     \STATE Break
%     \ENDIF
%     \STATE Choose level with greatest uncertainty
%     \STATE Calculate MA and KID for this perturbation level
%     \STATE Add $f(p)$ to the list of points

%    \UNTIL{$noChange$ is $true$}
% \end{algorithmic}
% \end{algorithm}

\section{Overall Performance on ACDC and IDD}
\label{appendix:realworld}
\begin{table}[H]
    \centering
    \scalebox{0.7}{
    \begin{tabular}{lrrrrrrr}
    \toprule 
    \phantom{a} & \multicolumn{3}{c}{Weather // ACDC} & \phantom{a} & \multicolumn{3}{c}{Domain // IDD} \\
    \cmidrule{2-4} \cmidrule{6-8}
    Method &  aAcc$\uparrow$ & mIoU$\uparrow$ & mAcc$\uparrow$ & \phantom{a} & aAcc$\uparrow$ & mIoU$\uparrow$ & mAcc$\uparrow$ \\
    \midrule
    Baseline & 76.31 & 35.48 & 47.36 & \phantom{a} & 85.82 & 38.44 & 59.14 \\
    AugMix & 79.57 & 40.90 & 52.74 & \phantom{a} & \textbf{86.52} & 40.50 & 62.43 \\
    AutoAugment & 70.29 & 39.31 & 54.18 & \phantom{a} & 85.79 & \textbf{40.74} & 62.24 \\
    RandAug & 78.46 & 39.07 & 52.32 & \phantom{a} & 85.54 & 38.99 & 59.82 \\
    TrivialAug & 75.50 & 38.56 & 53.62 & \phantom{a} & 85.23 & 39.61 & 61.04 \\
    IDBH & 78.65 & 41.67 & 53.65 & \phantom{a} & 86.49 & 40.48 & 61.74 \\
    VIPAug & 81.85 & 42.26 & 53.59 & \phantom{a} & 85.94 & 38.97 & 58.79 \\
    % \rowcolor{Gray} Ours & \textbf{82.01} & \textbf{46.54} & \textbf{57.82} & \phantom{a} & \textbf{86.87} & \textbf{42.52} & \textbf{64.82} \\
    \rowcolor{Gray} Ours & \textbf{80.16} & \textbf{45.45} & \textbf{57.58} & \phantom{a} & 85.76 & 40.33 & \textbf{63.03} \\
    % /fs/nexus-projects/robustness_datasets/iclr_rebuttal/ablation_eval/ours_segformer_cityscapes_full/acdc/20241123_050826/20241123_050826.json
    \bottomrule
    \end{tabular}
    }
    \caption{\textbf{Evaluation results on Unseen Real World Driving Datasets.} We conduct zero-shot evaluation of Cityscape models on both ACDC~\citep{acdc} and IDD~\citep{idd} datasets, which represent adverse weather and domain transfer to India respectively. Our method achieves clear improvements compared to other methods which require chained, more computationally expensive augmentations or external augmentation models in terms of generalization to real world scenarios, with relative mIoU improvement up to \textit{9.07\% on ACDC compared to the next-best, IDBH}.}
    \vspace{-1em}
    \label{tab:realworld}
    %(27.13 - 14.62) / 14.62 * 100 = 85.57
    %(33.09 - 21.80) / 21.80 * 100 = 51.79
\end{table}

\section{Downstream Finetuning with DinoV2}

\begin{table}[H]
    \centering
    \scalebox{1.0}{
    \begin{tabular}{lrrr}
    \toprule 
    \phantom{a} & \multicolumn{3}{c}{ViT+DinoV2} \\
    \cmidrule{2-4}
    Method & aAcc$\uparrow$ & mAcc$\uparrow$ & mIoU$\uparrow$ \\
    \midrule
    Baseline & 77.65 & 45.83 & 32.70\\
    Augmix & 79.99 & 51.63 & 41.38 \\
    AutoAugment & 81.18 & 55.93 & 43.65 \\
    RandAugment & 80.42 & 54.02 & 43.25 \\ 
    TrivialAugment & 82.56 & 54.27 & 43.58 \\
    IDBH & \textbf{84.45} & 60.22 & 48.69 \\
    % VIPAug & 85.87 & 64.66 & 52.32 \\
    \rowcolor{Gray} Ours & 84.13 & \textbf{62.92} & \textbf{49.82} \\
    % \rowcolor{Gray} Ours & \textbf{84.91} & \textbf{57.87} & \textbf{47.44}\\
    \bottomrule
    \end{tabular}
    }
    \caption{\textbf{Performance of Cityscapes models on \textit{unseen} ACDC weather evaluation set across different augmentation methods, when fine-tuned from DinoV2}~\citep{oquab2024dinov} with ViT~\citep{dosovitskiy2021an_vit} backbone.}
    \vspace{-1.5em}
\label{tab:foundation_finetuning}
\end{table}

\section{Downstream Finetuning with SAM}
% \begin{table}[H]
%     \centering
%     \begin{tabular}{lrrr}
%         \toprule
%         Method & aAcc$\uparrow$ & mAcc$\uparrow$ & mIoU$\uparrow$ \\
%         \midrule
%         Baseline & 84.93 & 62.84 & 52.20 \\
%         AugMix & 84.69 & 63.25 & 54.18\\
%         AutoAugment & 85.17 & 61.28 & 53.11 \\
%         RandAug & 85.16 & 59.33 & 51.95 \\
%         TrivialAugment & 84.87 & 59.92 & 50.58 \\
%         IDBH & 85.14 & 62.82 & 54.35 \\
%         VIPAug & 84.29 & 61.00 & 51.93 \\
%         Ours & \textbf{85.37} & \textbf{65.18} & \textbf{54.84} \\
%         \bottomrule
%     \end{tabular}
%     \caption{\textbf{Performance on ACDC when fine-tuning downstream segmentation with SAM.} We show additional comparisons when initialized with SAM weights, similarly to results in Table~\ref{tab:foundation_finetuning}.}
%     \label{tab:sam_ACDC_finetune}
%     \vspace*{-1em}
% \end{table}

\begin{table*}[th]
  \centering
  \scalebox{.9}{
  \begin{tabular}{lrrcrrcrrcrr}
    \toprule
      & \multicolumn{2}{c}{Fog} & \phantom{a} & \multicolumn{2}{c}{Rain} & \phantom{a} & \multicolumn{2}{c}{Night} & \phantom{a} & \multicolumn{2}{c}{Snow} \\
   \cmidrule{2-3} \cmidrule{5-6} \cmidrule{8-9} \cmidrule{11-12}
   Method & aAcc$\uparrow$ & mIoU$\uparrow$ & \phantom{a} & aAcc$\uparrow$ & mIoU$\uparrow$ & \phantom{a} & aAcc$\uparrow$ & mIoU$\uparrow$ & \phantom{a} & aAcc$\uparrow$ & mIoU$\uparrow$ \\
   \midrule
   Baseline & 93.96 & 68.98 & \phantom{a} & 91.59 & 58.53  & \phantom{a} & 65.73 & \textbf{29.41} & \phantom{a} & 90.54 & 56.27 \\ 
   AugMix & 94.38 & 70.49 & \phantom{a} & 92.18 & \textbf{60.66} & \phantom{a} & 62.47 & 25.35 & \phantom{a} & 90.46 & 58.60 \\ 
   AutoAugment & 94.16 & 69.18 & \phantom{a} & 92.57 & 56.63 & \phantom{a} & 63.66 & 26.69 & \phantom{a} & 90.98 & 58.19 \\ 
   RandAug & 93.92 & 67.67 & \phantom{a} & 91.70 & 55.30 & \phantom{a} & 65.19 & 24.00 & \phantom{a} & 90.47 & 56.70 \\ 
   TrivialAug & 93.71 & 66.45 & \phantom{a} & 91.06 & 53.09  & \phantom{a} & 60.57 & 20.25 & \phantom{a} & 90.56 & 54.67 \\
   IDBH & 94.12 & 70.61 & \phantom{a} & 92.06 & 55.71 & \phantom{a} & 65.25 & 28.22 & \phantom{a} & 90.98 & \textbf{59.67} \\
   VIPAug & 93.92 & 67.54 & \phantom{a} & 91.06 & 55.49 & \phantom{a} & 63.00 & 25.01 & \phantom{a} & 89.88 & 57.85 \\
   \rowcolor{Gray} Ours & \textbf{94.34} & \textbf{70.98} & \phantom{a} & \textbf{92.66} & 57.92 & \phantom{a} & \textbf{66.93} & 27.24 & \phantom{a} & \textbf{91.08} & 57.32 \\

    \bottomrule
  \end{tabular}}
  \caption{
    \textbf{Evaluation on ACDC adverse weather performance with SAM downstream finetuning.} 
    }
    \vspace{-1em}
  \label{tb:weather_evaluation_sam}
\end{table*}

% \section{Domain Adaptation Experiments}
% \begin{table}[H]
%     \centering
%     \begin{tabular}{lrrr}
%         \toprule
%         Method & aAcc$\uparrow$ & mIoU$\uparrow$ & mAcc$\uparrow$ \\
%         \midrule
%         Baseline & 94.19 & 66.67 & 75.28 \\
%         AugMix & 93.99 & 66.07 & 73.60\\
%         AutoAugment & 93.84 & 64.99 & 72.09 \\
%         RandAug & 92.60 & 59.37 & 66.03 \\
%         TrivialAugment & 93.55 & 65.03 & 71.71\\
%         IDBH & 93.62 & 65.49 & 74.96 \\
%         VIPAug & 92.98 & 63.67 & 70.86 \\
%         Ours & \textbf{93.88} & \textbf{68.03} & \textbf{75.96} \\
%         \bottomrule
%     \end{tabular}
%     \caption{\textbf{Domain adaptation experiments.} We show performance on the validation set of the ACDC Snow dataset, after training for 20k iterations on the ACDC Snow training set. All experiments are initialized with a Segformer-b0 model pre-trained on Cityscapes.}
%     \label{tab:sam_ACDC_finetune}
%     \vspace*{-1em}
% \end{table}

\section{Qualitative Results on Window Wiper Occlusion in ACDC}
\label{appendix:qualitative_windshield}

\begin{figure*}[H]
 \vspace{-0.5em}
    \centering
    \subfigure[Ground Truth.]{\includegraphics[width=0.4\linewidth]{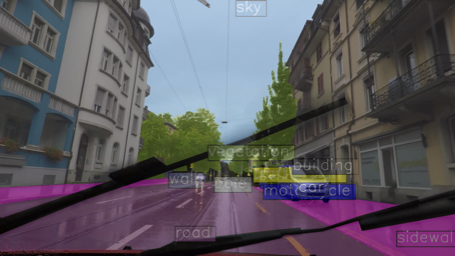}}\quad
    \subfigure[AutoAugment.]{\includegraphics[width=0.4\linewidth]{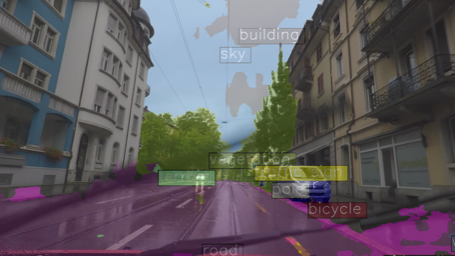}}\quad
    \subfigure[IDBH.]{\includegraphics[width=0.4\linewidth]{figures/qualitative_acdc/occlusion/idbh_rain_572_860.png}}\quad
    \subfigure[Ours.]
    {\includegraphics[width=0.4\linewidth]{figures/qualitative_acdc/occlusion/ours_572_860.png}}
\vspace{-1.3em}
    \caption{\textbf{Special case on ACDC prediction: windshield wiper occlusion.} We observe a special case of natural corruptions in rainy weather which cannot be directly simulated by the existing set of perturbations: physical occlusion by windshield wipers. \textbf{\textit{While IDBH involves random occlusion during training, ours does not. Our augmentation approach achieves comparable qualitative results with a smaller set of augmentations.}}}
    \vspace{-0.5em}
    \label{fig:acdc_fail}
\end{figure*}

\section{Detailed Experiment Hyperparameters}
\label{appendix:hyperparameter}
Let $AA =$ \{RandomCrop, Contrast, Equalize, Posterize, Rotate, Solarize, Shear X, Shear Y, Translate X, Translate Y, Color, Contrast, Brightness, Sharpness\} be the standard set of augmentations used with AutoAugment.

\begin{table}[H]
  \centering
  \scalebox{.8}{
  \begin{tabular}{ccccp{3cm}cc}
    \toprule
    Method & Max Iters & LR & Optimizer & Augmentations & Batch Size & Backbone \\
    \midrule 
    Baseline & 160,000 & 6e-05 & AdamW & RandomCrop & 1 & SegFormer-b0 \\
    Augmix & 160,000 & 6e-05 & AdamW & AA & 1 & SegFormer-b0 \\
    AutoAugment & 160,000 & 6e-05 & AdamW & AA & 1 & SegFormer-b0 \\
    RandAug & 160,000 & 6e-05 & AdamW & AA & 1 & SegFormer-b0 \\
    TrivialAug & 160,000 & 6e-05 & AdamW & AA & 1 & SegFormer-b0 \\
    IDBH & 160,000 & 6e-05 & AdamW & AA $\cup$ \{\textbf{RandomFlip}, \textbf{RandomErasing}\}& 1 & SegFormer-b0 \\
    Ours & 160,000 & 6e-05 & AdamW & AA & 1 & SegFormer-b0 \\
    $r_v=1600$; \\
    $r_{SA}=9600$; \\
    Ours & 160,000 & 6e-05 & AdamW & AA & 1 & SegFormer-b0 \\
    $Warmup=6400$ \\
    \bottomrule
  \end{tabular}}
 \caption{
\textbf{Experiment hyperparameters for Table~\ref{tab:realworld} and Table~\ref{tb:dataset_experiment_comparison_small} .} All experiments are trained under similar hyperparameter settings, with each evaluation conducted on the \textit{highest-performing mIoU checkpoint}. In comparisons, we prioritize official implementations released by authors and avoid re-implementations. Additionally, most comparisons use the same set of augmentations to ours, with the exception of IDBH~\cite{li2023data}, whose original implementation includes RandomFlip and RandomErasing. For all experiments, we use the SegFormer-b0 backbone~\cite{segformer}, which is a recent state-of-the-art segmentation-specialized architecture.
}
\vspace{-1em}
  \label{tb:hyperparameter_table}
\end{table}

\section{Qualitative Results on Synapse}
\begin{figure}[H]
  \centering
  \subfigure[Ground truth.]{\includegraphics[width=0.23\linewidth]{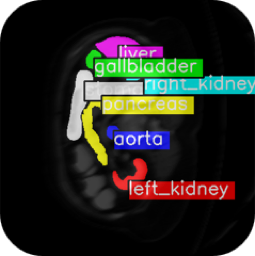}}\quad
  \subfigure[TrivialAugment Prediction.]
  {\includegraphics[width=0.23\linewidth]{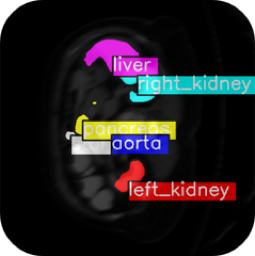}}\quad
  \subfigure[IDBH Prediction.]
  {\includegraphics[width=0.23\linewidth]{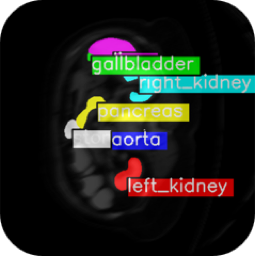}}
  \subfigure[Our Prediction.]
  {\includegraphics[width=0.23\linewidth]{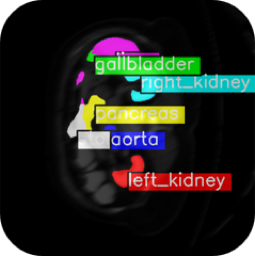}}
  \label{fig:synapse_qualitative}
  \vspace*{-1em}
  \caption{\textbf{Qualitative evaluation on multi-organ segmentation with motion blur corruption.} We show predictions on a motion-blurred sample from the Synapse~\citep{synapse} dataset for TrivialAugment (b), IDBH (c), and Our method (d), against the ground truth (a). Our method is able to segment \textcolor{cyan}{right} and \textcolor{red}{left} kidneys, \textcolor{hotpink}{liver}, and \textcolor{navyblue}{aorta} accurately. In contrast, the TrivialAugment prediction is unable to distinguish both kidneys.}
  \vspace*{-1em}
\end{figure}

\section{Qualitative Results on Rainy Data}
\label{appendix:rainy_data}
\begin{figure}[H]
    \centering
    \subfigure[AutoAugment Prediction.]{\includegraphics[width=0.30\linewidth]{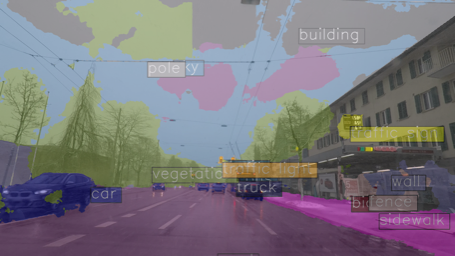}}\quad
    \subfigure[IDBH Prediction.]
    {\includegraphics[width=0.30\linewidth]{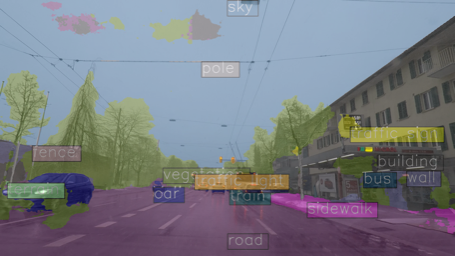}}\quad
    \subfigure[Our Prediction.]
      {\includegraphics[width=0.30\linewidth]{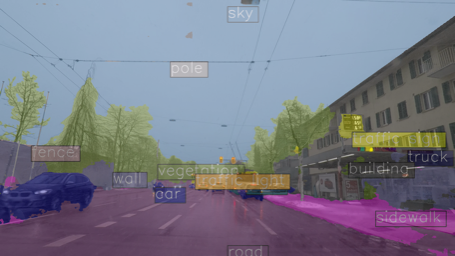}}
    \caption{\textbf{Qualitative comparison on snowy urban driving sample between AutoAugment~\cite{cubuk_2020}, IDBH~\cite{li2023data}, and Ours.} 
    In this example, each method (AutoAugment, IDBH, Ours) is trained on clean Cityscapes data representing sunny weather, then evaluated on adverse weather samples. Despite not having rainy data in the training set, our method is able to segment the driving noticeably clearer than other methods. In particular, other methods consistently struggle to segment the \textcolor{blue}{vehicle} confidently. }
    \label{fig:acdc_rainy}
\end{figure}

\section{Special Case: Windshield Wiper Occlusion}
\label{appendix:failure_acdc}
\begin{figure}[H]
    \centering
    \subfigure[Ground Truth.]{\includegraphics[width=0.23\linewidth]{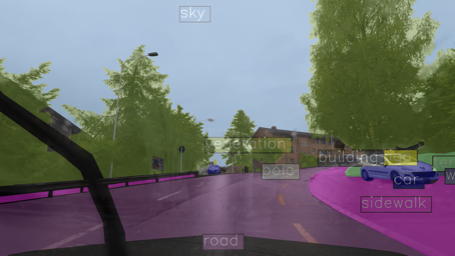}}\quad
    \subfigure[AutoAugment.]{\includegraphics[width=0.23\linewidth]{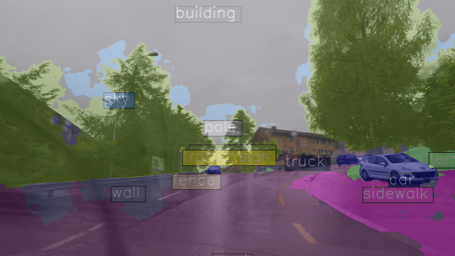}}\quad
    \subfigure[IDBH.]{\includegraphics[width=0.23\linewidth]{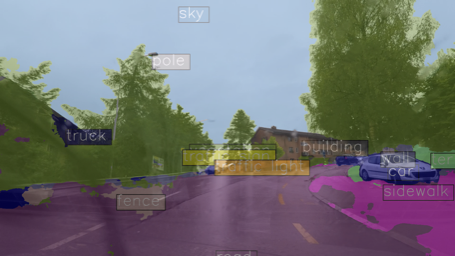}}\quad
    \subfigure[Ours.]
    {\includegraphics[width=0.23\linewidth]{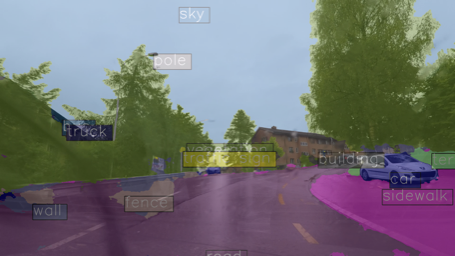}}
    \subfigure[Ground Truth.]{\includegraphics[width=0.23\linewidth]{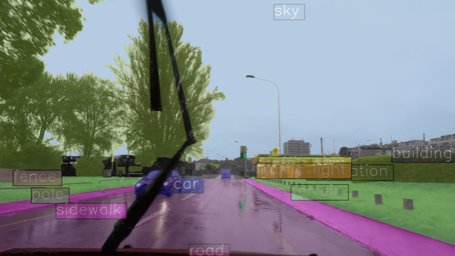}}\quad
    \subfigure[AutoAugment.]{\includegraphics[width=0.23\linewidth]{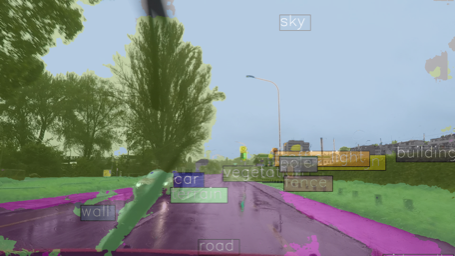}}\quad
    \subfigure[IDBH.]{\includegraphics[width=0.23\linewidth]{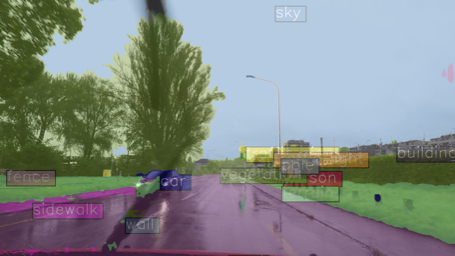}}\quad
    \subfigure[Ours.]
    {\includegraphics[width=0.23\linewidth]{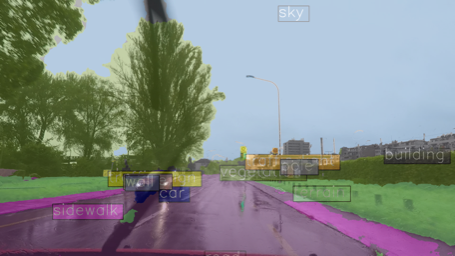}}
    \subfigure[Ground Truth.]{\includegraphics[width=0.23\linewidth]{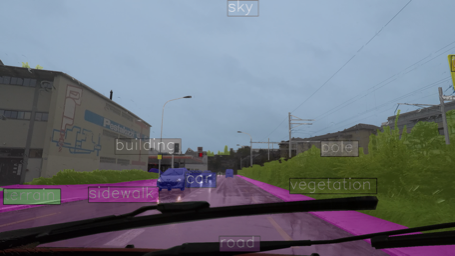}}\quad
    \subfigure[AutoAugment.]{\includegraphics[width=0.23\linewidth]{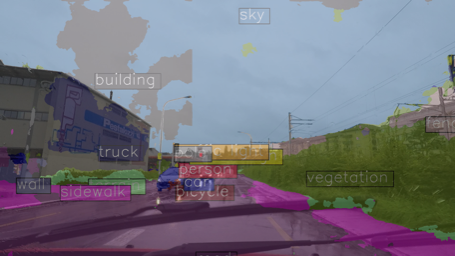}}\quad
    \subfigure[IDBH.]{\includegraphics[width=0.23\linewidth]{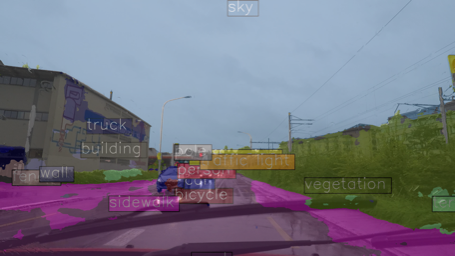}}\quad
    \subfigure[Ours.]
    {\includegraphics[width=0.23\linewidth]{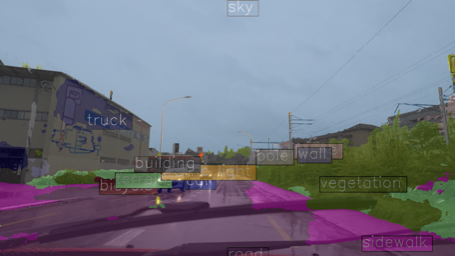}}
    \subfigure[Ground Truth.]{\includegraphics[width=0.23\linewidth]{figures/qualitative_acdc/occlusion/gt_572_860.png}}\quad
    \subfigure[AutoAugment.]{\includegraphics[width=0.23\linewidth]{figures/qualitative_acdc/occlusion/aa_rain_572_860.png}}\quad
    \subfigure[IDBH.]{\includegraphics[width=0.23\linewidth]{figures/qualitative_acdc/occlusion/idbh_rain_572_860.png}}\quad
    \subfigure[Ours.]
    {\includegraphics[width=0.23\linewidth]{figures/qualitative_acdc/occlusion/ours_572_860.png}}
    \caption{\textbf{More examples of special case on ACDC prediction: windshield wiper occlusion.}}
    \label{fig:acdc_rain_additional}
\end{figure}
% \subsection{Definition of Sensitivity Analysis}

% $$sensitivity = \frac{\partial MA(p)}{\partial KID(p)}$$

% Define $p_0$ and $p_n$ as the minimum and maximum perturbation levels respectively.

% $$Q = \argmax_{\substack{\{p_1, \ldots, p_{n - 1}\} \subset P \\ p_1 < \cdots < p_{n - 1}}} \, \min_{1 \leq i \leq n - 1} \frac{|MA(p_{i-1}) - MA(p_{i})|}{MA(p_n)} - \frac{|KID(p_i) - KID(p_{i - 1})|}{{KID(p_n)}} + 2\left(\frac{p_i - p_{i - 1}}{p_n}\right)$$

% Define
% $$f(p) = \frac{MA(p)}{MA(p_n)} - \frac{KID(p)}{KID(p_n)} + 2\left(\frac{p}{p_n}\right).$$

% Then,
% $$Q \approx \argmax_{\substack{\{p_1, \ldots, p_{n - 1}\} \subset P \\ p_1 < \cdots < p_{n - 1}}} \, \min_{1 \leq i \leq n - 1} f(p_i) - f(p_{i - 1})$$

% $$\sum_{i = 1}^n f(p_i) - f(p_{i - 1}) = f(p_n) - f(p_0) = 2$$

% $$Q \approx \left\{p_i \in P ~\Big\vert~ f(p_i) = \frac{2i}{n}, \, i = 1, \ldots, n\right\}$$

\subsection{Details on Basis Augmentations}
\label{sec:basis_perturbation}
% insert stuff here about basis perturbations, previous work, introduce notion of objective function 

% \subsection{Problem Definition}

% \subsection{Basis Perturbations}
Previous work in robustification showed that learning with a set of ``basis perturbations'' (BP) significantly improved zero-shot evaluation against unseen corruptions ~\cite{yu_shen} for image classification and regression tasks, such as vehicle steering prediction. 
The intuition behind basis perturbations is that the composition of such perturbations spans a much larger space of perturbations than may be observed in natural corruptions; observed zero-shot performance boosts on unseen corruptions subsequently might be attributed to learning a model which is robust to basis perturbations.
In our method, we extend this concept and introduce a more generalized and larger set of basis perturbations in sensitivity analysis to determine the most productive augmentation during training.

Let $D = \{Positive, Negative\}$ describe the set of augmentations applied in either a positive (lighter) direction or negative (darker) to either one channel of an image or a parameter of an affine transformation applied to an image. 

Let $P = \{R, G, B, H, S, V\}$ describe the set of channels in RGB and HSV color spaces which may be perturbed; in other words, these augmentations are \textit{photometric}.

Then, let $G = \{ShearX, ShearY, TranslateX, TranslateY, Rotate\}$ denote affine, or \textit{geometric}, transformations which are parameterized by a magnitude value. 

Finally, let $Z = \{Noise, Blur\}$ be the set of augmentations not applied along channel dimensions. Specifically, we use Gaussian Noise and Gaussian Blur.

Thus, the set of all basis augmentations $A_B$ used in robustification is $A_B = \{D \times P + G + Z\}$.

To compute lighter or darker channel augmentations of RGB or HSV channels, we use linear scaling. Let the range of a channel be $[v_{\min}, v_{\max}]$. For lighter channel augmentations, we transform the channel values $v_C$ by an intensity factor $\alpha$ like so: 
$$v_C' = \alpha v_{\max} + (1-\alpha) \cdot v_C$$
Likewise, for darker channel augmentations, the transformation can be described like so: 
$$v_C' = \alpha v_{\min} + (1-\alpha) \cdot v_C$$

The default values are $v_{\min} = 0$ and $v_{\max} = 255$. For $H$ channel augmentations, we set the maximum channel values to be $180$. For $V$ channel augmentations, we set the minimum channel values to be $10$ to exclude completely dark images.

Affine transformations can be represented as a $3 \times 3$ matrix, which, when multiplied with a 2-dimensional image, produces a geometrically distorted version of that image. 
Affine transformation matrices are typically structured in the form: 

$$
M = 
\begin{bmatrix}
    1 & Shear_X & T_x \\
    Shear_Y & 1 & T_y \\
    0 & 0 & 1 \\
\end{bmatrix}
$$

\noindent
for shear and translation transformations. For rotations where the center of the image is fixed as the origin point $(0,0)$, the transformation matrix is defined as: 

$$
M_{rot} = 
\begin{bmatrix}
    cos\theta & -sin\theta & 0 \\
    sin\theta & cos\theta & 0 \\
    0 & 0 & 1 \\
\end{bmatrix}
$$

To account for padded values in images after affine transformations, we zoom in images to the largest rectangle such that padded pixels are cropped out. 

All augmentations are parameterized by a magnitude value ranging from 0 to 1. 
A magnitude value of 1 corresponds to the most severe augmentation value. More details on exact parameter value ranges can be found in the appendix. 
Conversely, a magnitude value of 0 produces no changes to the original image, and can be considered an identity function.
We account for the symmetry of these augmentation transformations by considering both positive values and negative values as separate augmentations. 
The fast adaptive sensitivity analysis algorithm introduced in the next section relies on the property that increasing magnitude corresponds to increasing ``distance" between images. Thus, augmentations cannot simply span the value ranges -1 to 1, and we separate them instead to different augmentations (positive and negative).

We apply these augmentations on-the-fly in online learning rather than generating samples offline. 
Doing so greatly reduces the offline storage requirement by one order of magnitude. 
Suppose $L$ intensity levels are sampled for each basis augmentation. Then, offline generation of perturbed data requires up to $L \times 2 \times (|P| + |G|) + 2 = 24L$ additional copies of the original clean dataset.
{\em With online generation, we avoid offline dataset generation entirely} and only need the original clean dataset to be stored, similar to standard vanilla learning.

\begin{figure}[h!]
    \centering
    \vspace*{-0.5em}
    \includegraphics[width=0.95\textwidth]{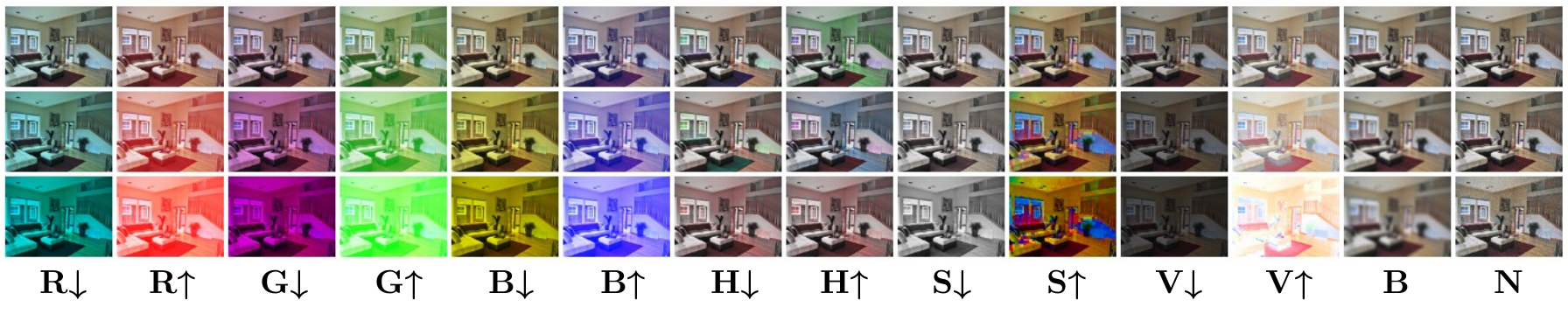}
    \vspace{-1.5em}
    \caption{\textbf{Visualization of each photometric augmentation transformation} on a bedroom image. Up $\uparrow$ indicates the ``lighter", positive direction and $\downarrow$ indicates the ``darker", negative direction. ``B" and ``N" indicate blur and noise, respectively.}
    % \vspace{-1em}
    \label{fig:perturbation_visualization_photometric}
\end{figure}

\begin{figure}[h!]
    \centering
    \includegraphics[width=0.975\textwidth]{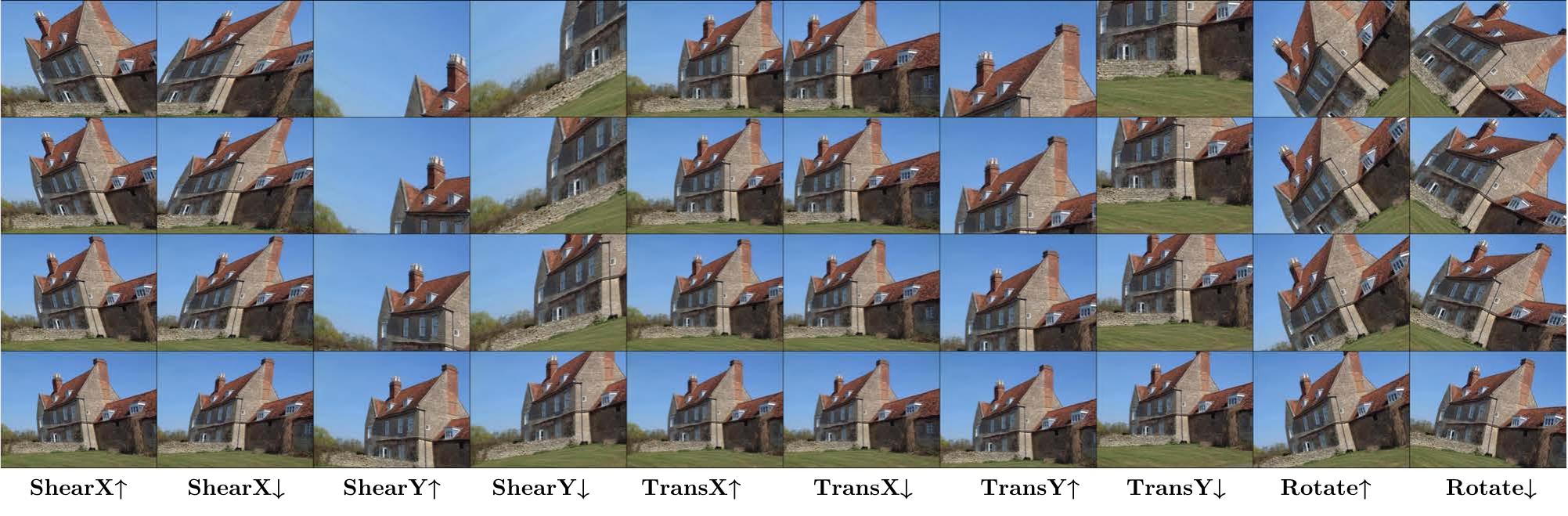}
    \vspace{-1em}
    \caption{\textbf{Visualization of various geometric augmentations} applied to a sample image of a house. We use the following geometric transformations in our sensitivity analysis scheme, which are also analogous to the set of transformations used by other methods~\cite{autoaugment, yu2022deepaa}. Up arrows indicate augmentation in the \textit{positive}, or left, direction, while down arrows indicate augmentation in the \textit{negative}, or right, direction.}
    % \vspace{-1.5em}
    \label{fig:perturbation_visualization_geometric}
\end{figure}

% \subsection{Additional Photometric Augmentations}
\begin{figure}[H]
    \centering
    \includegraphics[width=0.975\textwidth]{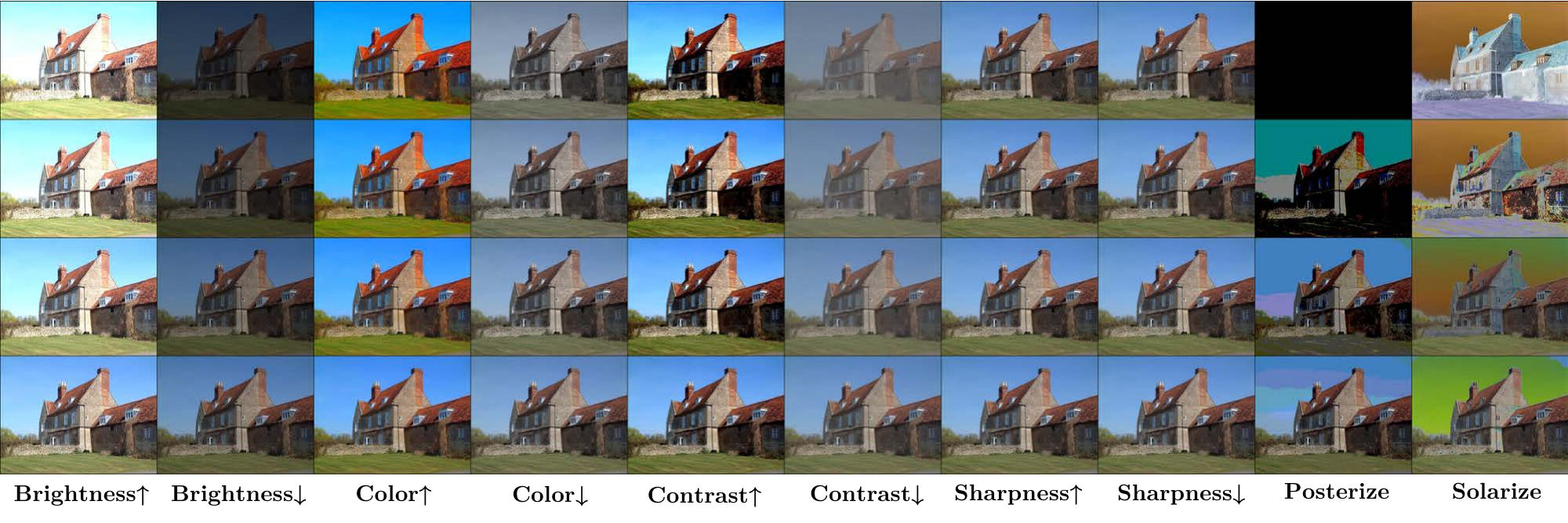}
    % \vspace{-2em}
    \caption{\textbf{Additional augmentation types used in sensitivity analysis, which are used in other methods such as AutoAugment. } While these photometric tranformations are used in other methods, the transformations also overlap with the photometric transformations shown in Figure~\ref{fig:perturbation_visualization_photometric}, namely HSV perturbations. However, we still conduct sensitivity analysis evaluation on these transformations for completion.}
    % \vspace{-1em}
    \label{fig:perturbation_visualization_photometric2}
    \vspace{1em}
\end{figure}

% \subsection{Qualitative Results on Snowy Weather}
% \begin{figure}[H]
%     \centering
%     \subfigure[RandAugment Prediction.]{\includegraphics[width=0.30\linewidth]{figures/qualitative_acdc/rand_snow_2.png}}\quad
%     \subfigure[TrivialAugment Prediction.]
%     {\includegraphics[width=0.30\linewidth]{figures/qualitative_acdc/trivial_snow_2.png}}\quad
%     \subfigure[Our Prediction.]
%       {\includegraphics[width=0.30\linewidth]{figures/qualitative_acdc/ours_snow_2.png}}
%     \caption{\textbf{Qualitative comparison on snowy example between RandAugment, TrivialAugment, and Ours.} Most noticeable failure in other methods lies within the sky area and neutral zones. Objects are classified similarly amongst methods, but a different sky texture or coloration may cause confusion with buildings, which may also cover large areas in similar areas of images.}
%     \label{fig:acdc_snowy}
% \end{figure}

\newpage
\section{AdvSteer Benchmark Examples}
\label{appendix:advsteer_examples}
\begin{figure}[H]
    \centering
    \includegraphics[width=0.8\linewidth]{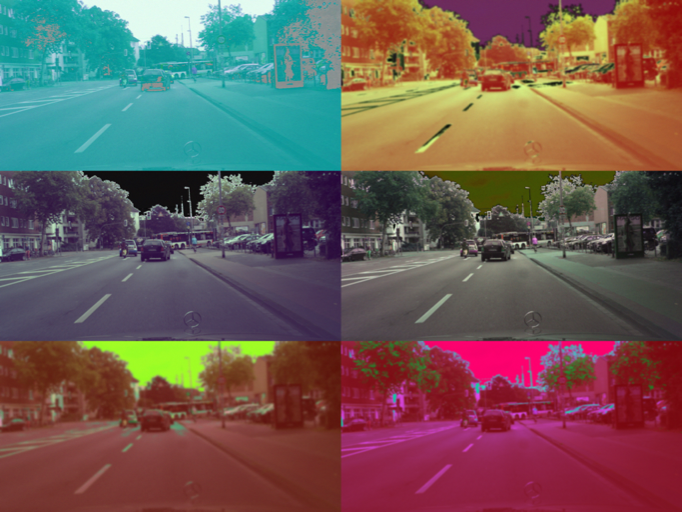}
    \caption{AdvSteer benchmark examples.}
    \label{fig:advsteer_examples}
\end{figure}
\newpage

\section{Clean Performance on Different Backbones}
\label{appendix:pspnet_segformer_comparison}
\begin{table}[H]
    \centering
    \scalebox{0.8}{
    \begin{tabular}{lrrrrrrr}
    \toprule 
    \phantom{a} & \multicolumn{3}{c}{PSPNet~\cite{zhao_2017}} & \phantom{a} & \multicolumn{3}{c}{SegFormer~\cite{segformer}} \\
    \cmidrule{2-4} \cmidrule{6-8}
    Method &  aAcc$\uparrow$ & mAcc$\uparrow$ & mIoU$\uparrow$ & \phantom{a} & aAcc$\uparrow$ & mAcc$\uparrow$ & mIoU$\uparrow$ \\
    \midrule
    Baseline & 63.770 & 48.695 & 35.715 & \phantom{a} & 86.825 & 57.280 & 48.365 \\
    Augmix & 94.770 & 74.400 & 66.740 & \phantom{a} & 95.520 & 81.430 & 73.390 \\
    AutoAugment & \textbf{95.130} & 77.210 & 69.630 & \phantom{a} & 95.550 & 81.390 & 73.820 \\
    RandAugment & 95.060 & 76.770 & 69.360 & \phantom{a} & 95.610 & 82.390 & 74.560 \\ 
    TrivialAugment & 95.090 & 75.930 & 68.620 & \phantom{a} & 95.640 & 83.210 & 75.130 \\
    % SOTA & 94.910 & 77.270 & 69.860  & \phantom{a} & 95.850 & 83.980 & 76.300 \\
    \rowcolor{Gray} Ours & 95.100 & \textbf{79.320} & \textbf{71.840} & \phantom{a} & \textbf{95.880} & \textbf{84.070} & \textbf{76.330} \\
    \bottomrule
    \end{tabular}
    }
    \caption{\textbf{Comparison of clean evaluation performance across different augmentation methods on Cityscapes. } We evaluated our sensitivity-informed augmentation method against popular benchmarks on PSPNet and SegFormer. The baseline represents training with no augmentations.}
    \vspace{-1em}
    \label{tab:method_experiment_small_architecture}
\end{table}

\section{Results on CUB Dataset for Classification}
\label{appendix:classification_results}
\begin{table}[h!]
    \centering
    \scalebox{0.8}{
    \begin{tabular}{lrrrr}
    \toprule 
    \phantom{a} & \multicolumn{4}{c}{InceptionV3} \\
    \cmidrule{2-5}
    Method & Clean & Basis Aug & AdvSteer & IN-C \\
    \midrule
    Baseline & 41.647  &  15.965  &  3.679 &  20.501 \\
    Augmix & 35.865  &  15.274  &  4.810 &  20.394 \\
    AutoAugment & 16.793   &  7.219  &  2.575   & 8.158 \\
    TrivialAugment & 33.914  &  13.338  &  4.229 &  17.586 \\
    RandAugment & 36.624  &  15.466  &  4.821 &  19.345 \\
    \rowcolor{Gray} Ours & \textbf{47.670}  &  \textbf{18.122}  &  \textbf{5.276}  & \textbf{21.842} \\
    \bottomrule
    \end{tabular}
    }
    \caption{Performance on CUB~\citep{WahCUB_200_2011} dataset with InceptionV3~\citep{inceptionv3} backbone.}
    \label{tab:cub_classification_experiment}
\end{table}

% \subsection{Fast Sensitivity Analysis Algorithm}
% \label{appendix:algorithm_sa}
% \input{sections/methodology_sections/algorithm_SA}
% \newpage

% \vspace*{3em}
\section{Fast Sensitivity Analysis Illustration}
\begin{figure}[ht!]
    \centering
    \includegraphics[width=0.8\textwidth]{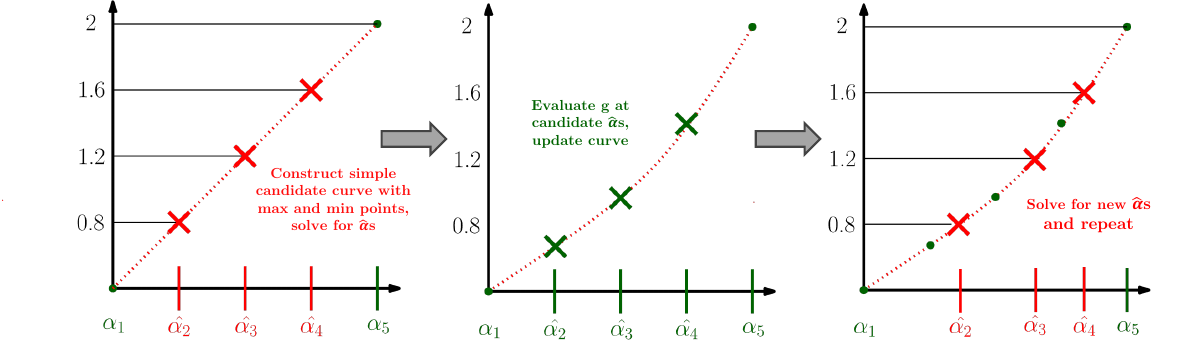}
    \caption{\textbf{Illustration of fast sensitivity analysis.} Each iteration of the fast sensitivity can be intuitively visualized. Since we can assume general monotonicity of the curve, we first initialize a candidate curve (a line in the first iteration). We solve for the candidate perturbation levels $\hat{\alpha}$ based on the solution in Equation~\ref{eq:q_solution}. In the next step (middle), we evaluate the candidate level with the greatest uncertainty and adjust the candidate curve, the dotted red line, using PCHIP on the evaluated levels, which are guaranteed to be true points along the function $g$ from Equation~\ref{eq:g_function}. In the next step (right), we use the new curve and solve for new candidate levels, repeating the process in the previous two steps until the maximum uncertainty of any candidate level values falls below a threshold of $0.05$.}
    \label{fig:sa_illustration}
\end{figure}
\newpage

\section{Sensitivity Analysis Computed Curve Comparison}
\begin{table}[H]
  \centering
  \scalebox{.9}{
  \begin{tabular}{ccrrrr}
    \toprule
    Perturb & Method & $p_1$ & $p_2$ & $p_3$ & $p_4$ \\
    \midrule
       \multirow{2}{*}{$R_{\uparrow}$} & Baseline & 0.100 & 0.300 & 0.500 & 0.700 \\
       & Adaptive & 0.149 & 0.253 & 0.399 & 0.604 \\
       \multirow{2}{*}{$G_{\uparrow}$} & Baseline & 0.100 & 0.200 & 0.400 & 0.600 \\
       & Adaptive & 0.103 & 0.204 & 0.395 & 0.619 \\
       \multirow{2}{*}{$B_{\uparrow}$} & Baseline & 0.200 & 0.300 & 0.500 & 0.700 \\
       & Adaptive & 0.146 & 0.328 & 0.551 & 0.788 \\
       \midrule
       \multirow{2}{*}{$R_{\downarrow}$} & Baseline & 0.200 & 0.400 & 0.600 & 0.800 \\
       & Adaptive & 0.225 & 0.503 & 0.625 & 0.803 \\
       \multirow{2}{*}{$G_{\downarrow}$} & Baseline & 0.200 & 0.400 & 0.600 & 0.800  \\
       & Adaptive & 0.256 & 0.447 & 0.607 & 0.812 \\
       \multirow{2}{*}{$B_{\downarrow}$} & Baseline & 0.200 & 0.500 & 0.700 & 0.800 \\
       & Adaptive & 0.231 & 0.450 & 0.594 & 0.730 \\
       \midrule
       \multirow{2}{*}{$H_{\uparrow}$} & Baseline & 0.100 & 0.300 & 0.400 & 0.900 \\
       & Adaptive & 0.268 & 0.406 & 0.508 & 0.809 \\
       \multirow{2}{*}{$S_{\uparrow}$} & Baseline & 0.200 & 0.500 & 0.600 & 0.800 \\
       & Adaptive & 0.243 & 0.439 & 0.589 & 0.744 \\
       \multirow{2}{*}{$V_{\uparrow}$} & Baseline & 0.200 & 0.400 & 0.600 & 0.700 \\
       & Adaptive & 0.193 & 0.360 & 0.517 & 0.680 \\
       \midrule
       \multirow{2}{*}{$H_{\downarrow}$} & Baseline & 0.200 & 0.400 & 0.500 & 0.600 \\
       & Adaptive & 0.279 & 0.433 & 0.548 & 0.699 \\
       \multirow{2}{*}{$S_{\downarrow}$} & Baseline & 0.200 & 0.400 & 0.600 & 0.900 \\
       & Adaptive & 0.199 & 0.344 & 0.562 & 0.847 \\
       \multirow{2}{*}{$V_{\downarrow}$} & Baseline & 0.200 & 0.400 & 0.600 & 0.800 \\
       & Adaptive & 0.197 & 0.397 & 0.594 & 0.797 \\
       \midrule
       \multirow{2}{*}{$blur$} & Baseline & 9 & 19 & 25 & 35 \\
       & Adaptive & 9 & 17 & 23 & 31 \\
       \multirow{2}{*}{$noise$} & Baseline & 10 & 15 & 20 & 30 \\
       & Adaptive & 6.4 & 12.4 & 17.7 & 26.9 \\
    \bottomrule
  \end{tabular}}
%   \vspace*{-0.25em}
     \caption{
    \textbf{Comparison of computed perturbation levels using a baseline~\cite{yu_shen} sensitivity analysis method versus our adaptive method.} $p_5$ is $1$ for all RGB/HSV channels, $49$ for blur, and $50$ for noise. In previous work, each perturbation level is chosen from a certain number of sampled, discretized values. Additionally, these perturbed datasets are generated offline in an additional step before training. Our fast sensitivity analysis enables sensitivity analysis to be performed on the fly during training, and offers much more dynamic, accurate, and descriptive sensitivity curves. 
    }
  \label{tb:sensitivity_analysis_comparison}
\end{table}
\newpage 

\subsection{KID vs. FID Relative Error Comparison with Scaling Sample Sizes}
\begin{figure}[H]
    \centering
    \includegraphics[width=0.5\columnwidth]{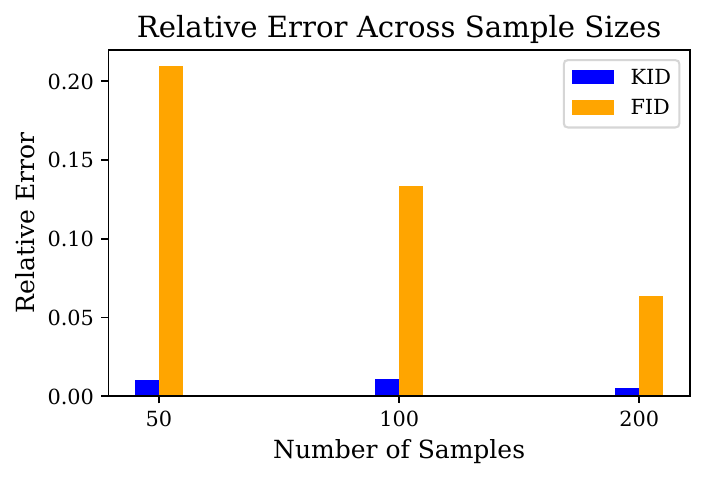}
    \caption{\textbf{Relative error of KID and FID over several sample sizes.} We plot the relative error of computed KID and FID values over several sample sizes, with the reference value being the computed value for each at 500 samples. From this, we can see that FID is significantly biased toward the number of samples used for evaluation. We can reduce the evaluation of KID values in sensitivity analysis by a notable fraction due to this property.}
    \label{fig:fidvskid}
    \vspace*{1em}
\end{figure}

\section{Train-time Evaluation on Perturbed Datasets}
\begin{figure}[H]
    \centering
    \includegraphics[width=\textwidth]{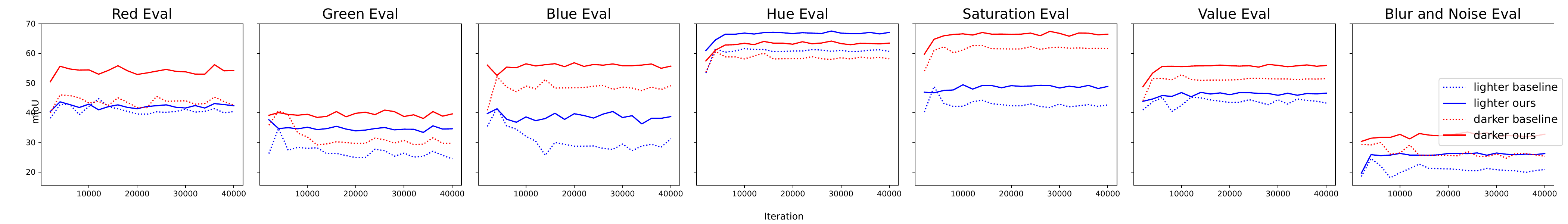}
    \caption{\textbf{Evaluation on perturbed test datasets over training iterations. } We show the evaluation on each perturbed dataset during training of our model and the baseline for VOC2012 dataset.}
    \label{fig:training_logs}
\end{figure}
\newpage 

\newpage 
\section{Adaptive Sensitivity Analysis with Different Number of Levels}
\begin{figure}[H]
    \centering
    \minipage{0.45\textwidth}
      \includegraphics[width=\linewidth]{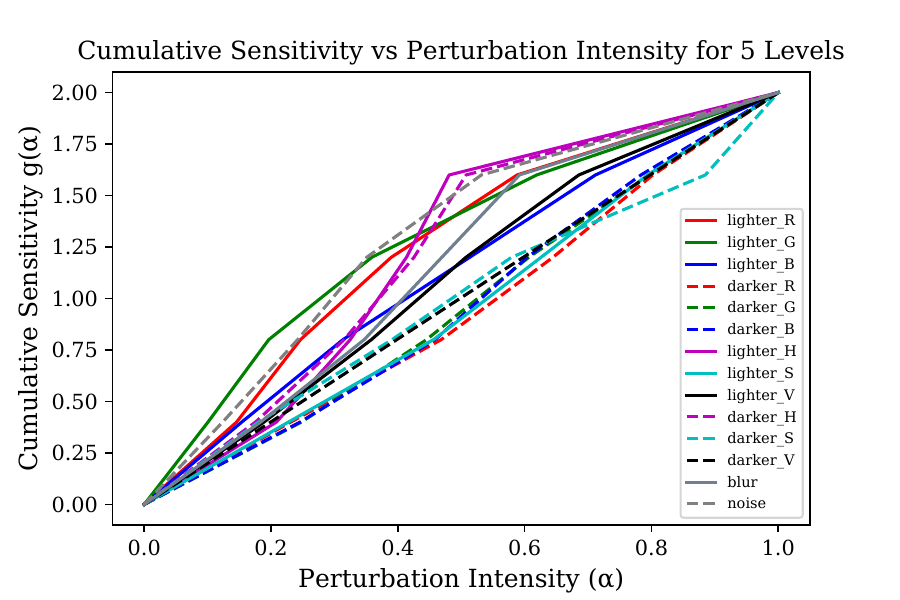}
    \endminipage
    \minipage{0.45\textwidth}
      \includegraphics[width=\linewidth]{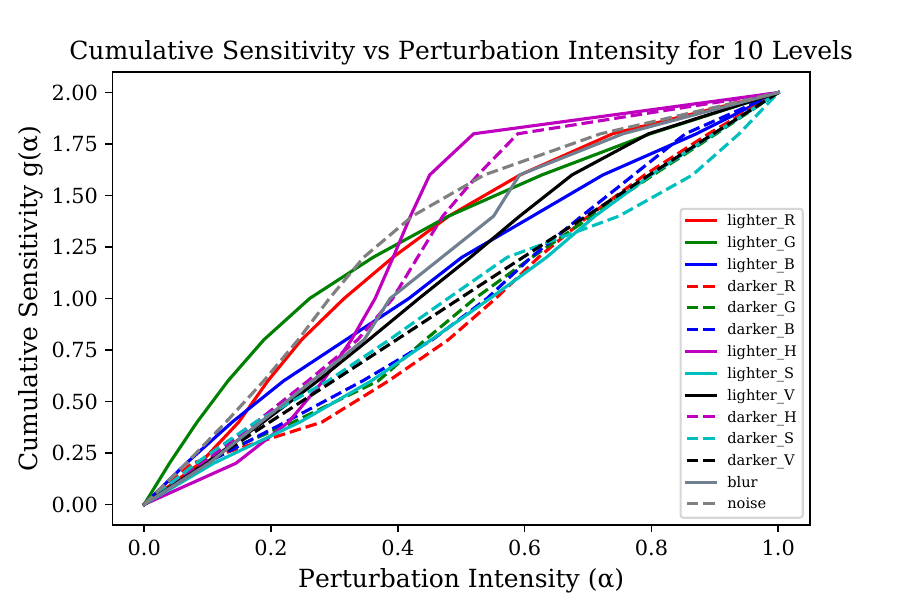}
    \endminipage\hfill
    \minipage{0.45\textwidth}
      \includegraphics[width=\linewidth]{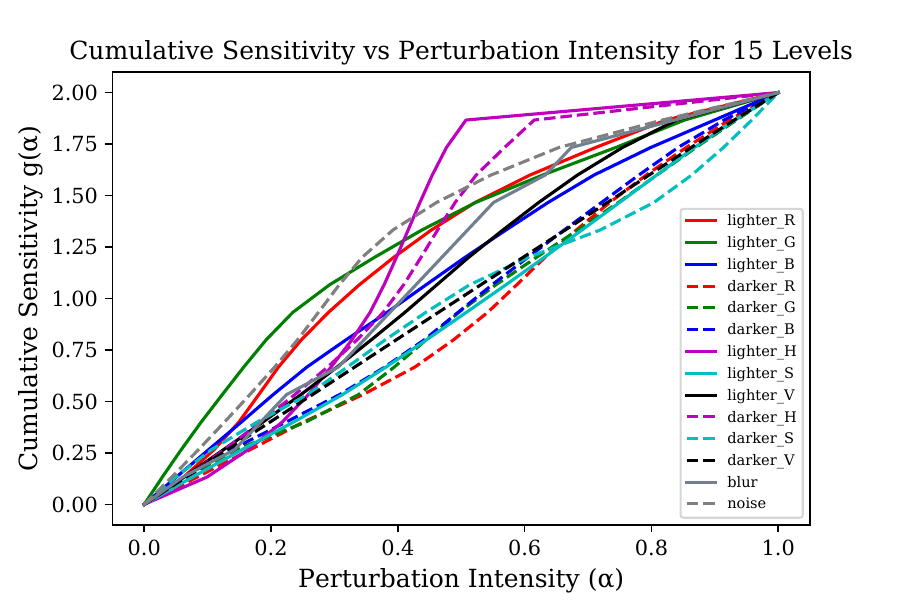}
    \endminipage
    \minipage{0.45\textwidth}
      \includegraphics[width=\linewidth]{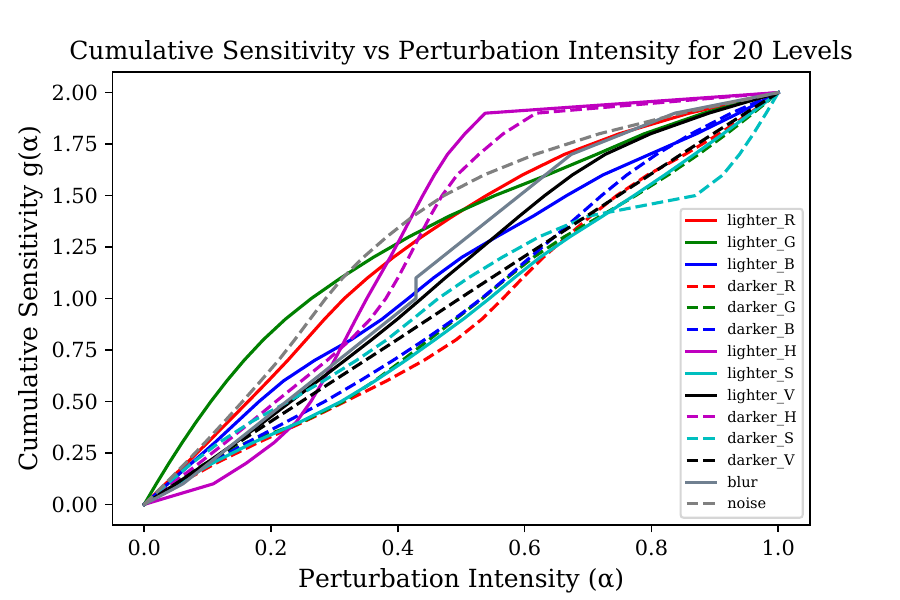}
    \endminipage
    \caption{\textbf{Visualization of cumulative sensitivity curve with varying number of levels $L$.} We visualize the cumulative sensitivity curve in Equation~\ref{eq:g_function} when computing for 5, 10, 15, and 20 levels. We find that even when we increase the number of levels, the curves remain \textit{approximately} the same. Thus, we use 5 levels in our implementation to reduce compute for the sensitivity analysis step. }
    \label{fig:sa_levels_testing}
    % \vspace*{1em}
\end{figure}

% \subsection{How Sensitivity Analysis Curve Changes over Training}
% \begin{figure}[htp]
%     \centering
%     \includegraphics[width=\linewidth]{figures/sa_curve_training.pdf}
%     \caption{\textbf{Cumulative sensitivity curves throughout training.} We visualize how the estimated cumulative sensitivity curve changes during augmented training. In this plot, the most recent curve is opaque, while others are decrease in opacity in order of recency.}
%     \label{fig:sa_curve_training}
%     \vspace*{-1em}
% \end{figure}

\end{document}